\documentclass{article}

\usepackage[preprint,nonatbib]{neurips_2024}

\usepackage[utf8]{inputenc} 
\usepackage[T1]{fontenc}    
\usepackage{hyperref}       
\hypersetup{
    colorlinks=true,        
    linkcolor=red,          
    urlcolor=magenta,       
    citecolor=Blue9, 
}
\usepackage{url}            
\usepackage{booktabs}       
\usepackage{amsfonts}       
\usepackage{nicefrac}       
\usepackage{microtype}      
\usepackage{xcolor}         

\definecolor{Blue9}{rgb}{0.098,0.3,0.9}

\usepackage{amsmath,amsfonts,bm}

\def\eqref#1{Eq.~(\ref{#1})}

\def\1{\bm{1}}

\newcommand{\ourmethod}{CARD\xspace}
\newcommand{\designer}{Coder\xspace}
\newcommand{\critic}{Evaluator\xspace}
\newcommand{\TPE}{TPE\xspace}
\newcommand{\MW}{Meta-World\xspace}
\newcommand{\MS}{ManiSkill2\xspace}
\newcommand{\LiftCube}{Lift Cube\xspace}
\newcommand{\PickCube}{Pick Cube\xspace}
\newcommand{\TurnFaucet}{Turn Faucet\xspace}
\newcommand{\OpenCabinetDrawer}{Open Cabinet Drawer\xspace}
\newcommand{\OpenCabinetDoor}{Open Cabinet Door\xspace}
\newcommand{\PushChair}{Push Chair\xspace}

\newcommand{\lr}{L2R\xspace}
\newcommand{\tr}{Text2Reward\xspace}
\newcommand{\eureka}{Eureka\xspace}

\newcommand{\PbRL}{Preference-based RL\xspace}
\newcommand{\pbRL}{preference-based RL\xspace}

\usepackage{amsmath}
\usepackage{amssymb}
\usepackage{mathtools}
\usepackage{amsthm}
\usepackage{mathtools} 

\theoremstyle{plain}
\newtheorem{theorem}{Theorem}[section]

\theoremstyle{definition}
\newtheorem{definition}[theorem]{Definition}

\theoremstyle{remark}

\usepackage{algorithm}
\usepackage{algorithmicx}
\usepackage{algpseudocode}

\usepackage{enumitem} 
\usepackage{subfig} 
\usepackage{multirow}

\usepackage{xspace}
\usepackage{wrapfig}  
\usepackage{makecell} 

\usepackage{graphicx} 
\usepackage{colortbl}
\usepackage{tcolorbox}

\graphicspath{{figs/}}

\usepackage{bbding}
\usepackage{verbatim}
\usepackage{fancyvrb}
\usepackage{listings}
\usepackage{threeparttable}

\setlist[itemize]{leftmargin=20pt}

\title{A Large Language Model-Driven Reward Design Framework via Dynamic Feedback for Reinforcement Learning}

\author{%
\hspace{-0.5em}
Shengjie Sun\textsuperscript{$1,$}\thanks{Equal contribution.} \hspace{0.005pt} \textsuperscript{$,$}\thanks{Work done during an internship at Tencent.} \ , Runze Liu\textsuperscript{$1,*$}, Jiafei Lyu\textsuperscript{$1$}, Jing-Wen Yang\textsuperscript{$2$}, Liangpeng Zhang\textsuperscript{$2$}, Xiu Li\textsuperscript{$1,$}\thanks{Corresponding author: Xiu Li~(li.xiu@sz.tsinghua.edu.cn)} \\
\textsuperscript{$1$}Tsinghua Shenzhen International Graduate School, Tsinghua University, \textsuperscript{$2$}Tencent IEG
}

\begin{document}

\maketitle

\begin{abstract}
Large Language Models (LLMs) have shown significant potential in designing reward functions for Reinforcement Learning (RL) tasks. However, obtaining high-quality reward code often involves human intervention, numerous LLM queries, or repetitive RL training. To address these issues, we propose CARD, a LLM-driven Reward Design framework that iteratively generates and improves reward function code. Specifically, CARD includes a Coder that generates and verifies the code, while a Evaluator provides dynamic feedback to guide the Coder in improving the code, eliminating the need for human feedback. In addition to process feedback and trajectory feedback, we introduce Trajectory Preference Evaluation (TPE), which evaluates the current reward function based on trajectory preferences. If the code fails the TPE, the Evaluator provides preference feedback, avoiding RL training at every iteration and making the reward function better aligned with the task objective. Empirical results on Meta-World and ManiSkill2 demonstrate that our method achieves an effective balance between task performance and token efficiency, outperforming or matching the baselines across all tasks. On 10 out of 12 tasks, CARD shows better or comparable performance to policies trained with expert-designed rewards, and our method even surpasses the oracle on 3 tasks.
\end{abstract}

\section{Introduction}

Reinforcement Learning (RL) has been successfully applied to various tasks with well-defined reward functions~\cite{mnih2015human,silver2016mastering,berner2019dota}. However, such reward functions do not exist for many real-world scenarios. A common approach is to manually design a reward function, known as reward engineering~\cite{laud2004theory,gupta2022unpacking}, but this requires extensive human knowledge and efforts. To tackle this problem, prior work has explored learning reward functions via inverse RL~\cite{ziebart2008maximum,wulfmeier2015maximum,finn2016guided,GAIL,AIRL} and \pbRL~\cite{christiano2017deep,ibarz2018reward,PEBBLE,Meta-Reward-Net}. However, inverse RL requires high-quality demonstrations and \pbRL often needs a large number of preference labels, both of which limit their use in practical applications.

Recently, Large Language Models (LLMs) have been demonstrated to be effective in generating reward function code for RL tasks~\cite{L2R,Text2Reward,Eureka,li2024auto}. However, the hallucination problem~\cite{rawte2023survey,huang2023survey,liu2024exploring} makes it difficult to generate an effective reward function with a single query. Existing methods require human feedback or incur high costs to improve the quality of generated reward functions. For example, \cite{L2R} relies on humans to manually design Motion Description templates and reward APIs, and the quality of the reward function is directly influenced by the accuracy of these designs. \cite{Text2Reward} requires humans to improve the reward function by analyzing agent trajectories, while \cite{li2024auto} leverages LLMs to analyze trajectories to refine the reward function. \cite{Eureka} identifies the most effective reward function using ground-truth rewards after repeated experiments, which significantly increases token consumption and RL training costs. Reducing human involvement, token consumption, and repetitive RL training while maintaining task performance remains a great challenge in LLM-based reward design.

\begin{figure*}[t]
\centering
\includegraphics[width=0.9\textwidth]{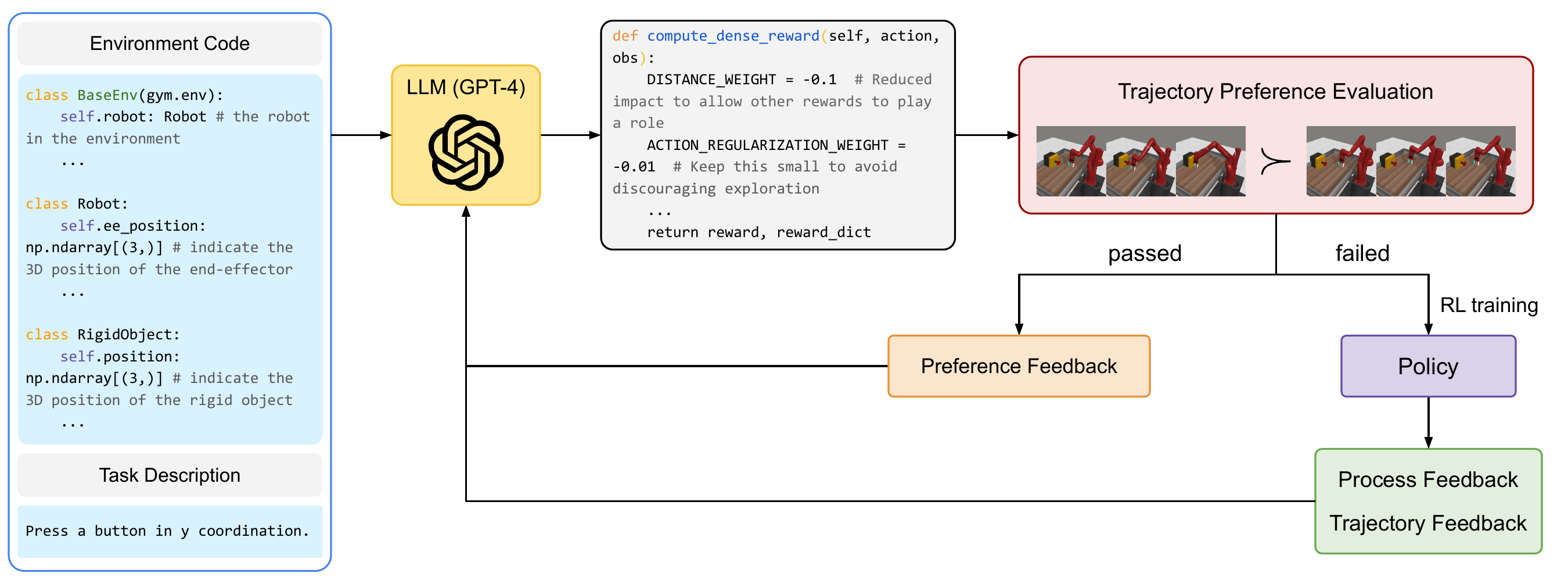}
\caption{\ourmethod includes a LLM-based \designer to generate reward function code and a \critic to evaluate the quality of the code. The \critic dynamically provides feedback to the \designer for reward function refinement.}
\label{fig:pipeline}
\end{figure*}

To address these issues, we propose a LLM-based \textbf{C}oder-Ev\textbf{A}luator \textbf{R}eward \textbf{D}esign framework, named \ourmethod. \ourmethod consists of a LLM-based \designer to generate reward function code and a \critic to evaluate the quality of the code, as illustrated in Figure~\ref{fig:pipeline}.
The \designer queries the LLM using environment description and task goals to generate initial reward function code, which is then verified for successful execution. The \critic dynamically provides feedback based on the quality of the reward function, without requiring additional LLM queries or human involvement. The \designer iteratively improves the reward function based on the provided feedback. We introduce Trajectory Preference Evaluation (\TPE), which evaluates the reward function with trajectory preferences, allowing the \critic to assess reward functions without additional RL training. If the reward function does not pass the \TPE, preference feedback is provided to improve the code, while process and trajectory feedback are provided otherwise. Specifically, process feedback records changes in parameters, such as trajectory return and sub-rewards during RL training. This allows the \designer to better understand the overall trends of the training process and evaluate the effectiveness of sub-reward components. Trajectory feedback provides parameter details for each step of successful and failed trajectories, enabling the \designer to compare the impact of sub-reward components on these trajectories. Preference feedback evaluates the reward function using trajectory pairs and preferences via \TPE without RL training results. The differences between \ourmethod and previous methods are listed in Table~\ref{tab:method_comparison}.
The contributions of this paper can be summarized as follows:
\begin{itemize}
    \item We propose a LLM-based reward design framework that generates and iteratively refines reward code without human feedback or parallel LLM queries.
    \item We propose Trajectory Preference Evaluation, enabling the \critic to dynamically provide feedback for reward function improvement, eliminating repetitive RL training.
    \item Empirical results on multiple tasks of \MW and \MS show that \ourmethod outperforms the baselines and even exceeds the human oracle.
\end{itemize}

\begin{table*}[!t]
\centering
\caption{Comparison of LLM-based reward design methods.}
\resizebox{1\linewidth}{!}{
\begin{tabular}{lcccccc}
\toprule
    \textbf{Method} & \begin{tabular}[c]{c}\textbf{Iterative}\\ \textbf{Improvement}\end{tabular} & \begin{tabular}[c]{c}\textbf{Human}\\ \textbf{Feedback}\end{tabular} & \begin{tabular}[c]{c}\textbf{Per-Iteration}\\ \textbf{RL Training}\end{tabular} & \begin{tabular}[c]{c}\textbf{Evaluator}\\ \textbf{Feedback}\end{tabular} & \begin{tabular}[c]{c}\textbf{\# of LLM Queries}\\ \textbf{Per Iteration}\end{tabular} & \begin{tabular}[c]{c}\textbf{\# of}\\ \textbf{Iterations}\end{tabular} \\
\midrule
    \lr~\cite{L2R} & \XSolidBrush & \XSolidBrush & \XSolidBrush & \XSolidBrush & $1$ & $0$ \\
    \tr\footnotemark[1]~\cite{Text2Reward} & \Checkmark & \Checkmark & \Checkmark & \XSolidBrush & $1$ & $0$ \\
    \eureka~\cite{Eureka} & \Checkmark & \XSolidBrush & \Checkmark & \XSolidBrush & $16$ & $5$ \\
    \ourmethod & \Checkmark & \XSolidBrush & \XSolidBrush & \Checkmark & $1$ & $2$ \\
\bottomrule
\end{tabular}
}
\label{tab:method_comparison}
\end{table*}

\footnotetext[1]{\tr supports both reward iterative improvement with additional human feedback and single-flow reward generation without human intervention.}
\setcounter{footnote}{1}

\section{Related Work}

\paragraph{Reward Design.}
Reward engineering remains a significant challenge in RL~\cite{laud2004theory,singh2009rewards,sutton2018reinforcement,gupta2022unpacking}.
The quality of the reward function is crucial for RL algorithms, especially for applications in real-world scenarios. Numerous research works investigate the construction of high-quality reward functions from diverse perspectives. In Imitation Learning (IL),~\cite{ziebart2008maximum,wulfmeier2015maximum,finn2016guided,GAIL,AIRL} extract a reward function from expert demonstrations via inverse RL.
In offline IL setting,~\cite{OTR} obtains rewards via optimal transport.~\cite{SEABO} utilizes a search-based method to design a reward function. These methods require expensive expert demonstrations. \PbRL learns from preference relations to obtain reward functions~\cite{christiano2017deep,ibarz2018reward,PEBBLE,Meta-Reward-Net,PEARL}. However, \pbRL often requires a large number of preference labels, which is expensive and time-consuming for human experts to provide.

\ourmethod differs from the above methods, as it does not require high-quality demonstration of preference data. Instead, \ourmethod leverages the impressive comprehension and generation capabilities of LLM to automatically generate and enhance reward functions without human knowledge.

\paragraph{Large Language Models for Reinforcement Learning.}
Many previous studies attempt to use LLM to assist in RL training. \cite{ma2022vip,fan2022minedojo,ma2023liv,du2023vision,karamcheti2023language,ELLM,kwon2023reward} utilize pretrained foundation models to generate reward signals for RL tasks. However, the agent needs a large number of samples from the environment during learning process. Frequent queries to the LLM not only decrease training efficiency but also lead to significant token consumption. Additionally, most of these methods generate scalar reward values, which are difficult to interpret and improve. Other approaches focus on solving robot control~\cite{liang2023code,huang2023instruct2act,wang2023demo2code} and goal planning~\cite{lin2023text2motion,ding2023task,xie2023translating,liu2023llm+,singh2023progprompt,wang2023voyager,silver2024generalized} problems by generating clear and structured program code. However, most of these works focus on executing robotic actions based on known motion primitives instead of learning low-level skills. Recent studies~\cite{L2R,Text2Reward,Eureka,li2024auto} address the challenge of robotic low-level manipulation in RL by creating reward functions using LLMs. However, these methods require human knowledge or lead to significant token consumption. In contrast, \ourmethod can generate reward function code and automatically provide feedback on its quality without human involvement or LLM analysis. Additionally, \ourmethod utilizes \TPE to eliminate unnecessary RL training, which further reduces the cost.

\section{Problem Setup}

We consider the standard RL setting, which is formulated as a Markov Decision Process (MDP). The MDP is defined by a tuple $(\mathcal{S}, \mathcal{A}, \mathcal{P}, \mathcal{R}, \gamma)$, where $\mathcal{S}$ is the state space, $\mathcal{A}$ is the action space, $\mathcal{P}$ is the transition dynamics, $\mathcal{R}$ is the reward function, and $\gamma \in [0,1)$ is the discount factor. The agent interacts with the environment at discrete time steps. At each time step $t$, the agent observes the state $s_t$, selects an action $a_t$ based on the policy $\pi$, and receives a reward $r_t$. The objective of the agent is to maximize the expected return $G = \sum_{t=0}^T \gamma^t r_t$, where $T$ is the horizon.

\section{Method}\label{sec:method}

In this section, we present \ourmethod, which consists of a \designer for generating reward function code and a \critic for evaluating its quality through an iterative process. \ourmethod involves three key steps: (1) Reward Design—the \designer generates the reward function code based on descriptions of the environment and task goals, followed by verification of its correctness; (2) Reward Introspection—the \critic provides various types of feedback on the quality and effectiveness of the trained policy, without relying on LLM queries or human intervention; and (3) Reward Improvement—the \designer refines the reward function code based on the feedback, forming an automatic refinement cycle. We first introduce the reward design process (Section~\ref{sec:reward_design}), followed by the reward introspection process (Section~\ref{sec:reward_introspection}), the reward improvement process (Section~\ref{sec:reward_improvement}), and finally, the practical implementation (Section~\ref{sec:practical_implementation}). The overall pipeline of \ourmethod is illustrated in Figure~\ref{fig:pipeline}, and the pseudo code for the algorithm is provided in Appendix~\ref{app:full_algorithm}.

\subsection{Reward Design}\label{sec:reward_design}

\paragraph{Code Generation.}

At this stage, the \designer queries the LLM to obtain the unchecked reward function code. However, directly generating domain-specific reward functions demands relevant prior knowledge, such as the configuration of the environment and the agent. Therefore, the \designer is provided with an environment description in Pythonic style, as shown in Figure~\ref{fig:pipeline}. Specifically, the environment abstraction shows callable variables, callable functions, and the inheritance relationship between different objects in the form of comments, static variable types, and Python classes.
Then, the \designer is provided with a task description and reward function template, querying the LLM to generate reward code that can guide the agent to complete the task. The \designer requires that the generated reward function not only returns a scalar reward signal but also specifies the value of each sub-reward~\cite{Eureka}. This guides the LLM to design reward code from the perspective of multiple reward components. The values of sub-rewards are used for \critic design feedback, detailed in Section~\ref{sec:reward_introspection}. We refer to the environment abstract description and task description of \MW and \MS in \cite{Text2Reward} and make
modifications to adapt them to our framework. For prompting details, please refer to Appendix~\ref{app:system_and_generation_prompt}.

\paragraph{Correctness Check.}

The reward function code generated by the LLM may occasionally contain syntax errors (e.g., usage of undefined variables) or runtime errors (e.g., variable type mismatch, matrix dimension mismatch), which can hinder downstream RL training. One solution is to provide error information as feedback to the LLM to fix the code~\cite{le2022coderl,olausson2023demystifying}, but this may lead the LLM to alter the original logic to correct the error. Another method is to generate multiple independent reward functions in parallel to ensure that at least one is correct~\cite{Eureka}. However, ensuring correct execution may require repeating the generation process multiple times, leading to unnecessary token usage. Unlike these methods, the \designer provides comments and type annotations for variables in the environment abstraction and reward function template to help the LLM clarify the environment settings. For unchecked reward functions, the \designer uses lightweight dynamic execution tests to check for syntax and runtime errors. If a test reports an error, the result is discarded, and the code is regenerated. The \designer will only re-initiate a query when the test fails, rather than generating a large amount of code at once. This ensures correct code is found without wasting tokens. The design details are in Appendix~\ref{app:implementation_details}. We compare the accuracy of reward design by different algorithms in Section~\ref{sec:execution_error_rate}.

\subsection{Reward Introspection}\label{sec:reward_introspection}

Due to the hallucination problem, the LLM may not always provide the optimal reward function in a single query. Previous methods either use LLM feedback~\cite{li2024auto} or human feedback~\cite{Text2Reward} to refine reward code, or select the best reward function from a set of independent experiments~\cite{Eureka}. To avoid human feedback or additional LLM queries, \ourmethod leverages \critic to automatically generate feedback through introspection. 
Specifically, the \critic generates three types of feedback: process feedback, trajectory feedback, and preference feedback, as illustrated in Figure~\ref{fig:pipeline}. Details of these feedback types are provided in Appendix~\ref{app:introspection_prompt}.

\subsubsection{Process Feedback}

Since the training curve reflects the quality of the reward function, we use the training results as feedback for evaluating the reward function. During RL training, the \critic evaluates the current policy by recording the average return, average trajectory length, and average success rate at each evaluation step, similar to~\cite{Eureka}. Instead of relying solely on the final result, these dense indicators help the \designer understand how the reward function influences the training process and identify areas for adjustment. Additionally, the \critic monitors the average cumulative value of each sub-reward along the trajectory, enabling specific adjustments to ineffective sub-rewards. The \critic then combines data, including average return, trajectory length, success rate, and the cumulative values of sub-rewards, into structured text in a natural language format as process feedback at regular intervals. Further details on the construction of process feedback can be found in Appendix~\ref{app:introspection_prompt}.

\subsubsection{Trajectory Feedback}

The process feedback provides detailed information about the training process. \ourmethod additionally incorporates feedback by analyzing sampled trajectories and such kind of comparison of reward values and states across trajectories is valuable for improving the reward function~\cite{li2024auto}. After RL training, \ourmethod rolls out trajectories using the trained policy and selects those with the highest and lowest returns to generate trajectory feedback. Specifically, the \critic gathers overall information about the trajectory, including return and success flags, while also sampling reward values, sub-reward values, and various observation parameters at intervals. This collected data is then integrated into natural language text as trajectory feedback. Since tasks with continuous state spaces may show minimal state changes per step, the sampling interval is treated as a hyperparameter. Details on the construction of trajectory feedback are provided in Appendix~\ref{app:introspection_prompt}. Unlike~\cite{li2024auto}, the \critic efficiently constructs trajectory feedback offline without relying on an LLM-based \textit{Trajectory Analyzer}, which improves the efficiency of reward introspection and reduces token costs.

\subsubsection{Trajectory Preference Evaluation and Preference Feedback}

Process feedback and trajectory feedback are both constructed based on RL training results, offering guidance for improving the reward function. Iteratively updating the reward function by training RL agents at each iteration is highly inefficient. In addition to repeated RL training, \eureka identifies the optimal reward function by comparing the trajectory return with the ground-truth reward. To address these issues, we introduce Trajectory Preference Evaluation (\TPE), which evaluates the reward function using trajectory preferences without requiring a ground-truth reward or repetitive RL training. The key insight behind \TPE is that some trajectories successfully achieve the goal, while others do not, naturally providing trajectory preference relations. \TPE offers a clear criterion for evaluating the quality of a reward function: if the return of successful trajectories exceeds that of unsuccessful ones when computed using the current reward function, then the reward function is considered order-preserving.

In many goal-conditioned environments, episodes are typically terminated upon task completion. As a result, successful trajectories tend to be shorter than unsuccessful ones. This can lead to longer, unsuccessful trajectories accumulating more rewards simply due to their length, even though they fail to complete the goal within a specified steps. To mitigate the effect of trajectory length, we compare the average per-step return rather than the cumulative return. The following provides a formal definition of order-preserving reward functions.

\begin{definition}

Given a set of trajectories $\tau = \{\tau_1, \tau_2, \dots, \tau_N\}$ where trajectories can have different lengths, a set $\tau^{+} \subseteq \tau$ containing successful trajectories, and a set $\tau^{-} \subseteq \tau$ containing unsuccessful trajectories, a reward function $r(s,a)$ designed by the \designer is order-preserving if the following condition holds:

\begin{equation}
\begin{aligned}
\min_{\tau_{i}^{+} \in \tau^{+}}\left(\frac{1}{|\tau_i^+|}\sum_{t=0}^{|\tau_i^+|} \gamma^{t} r(s_{it}^+, a_{it}^+)\right) > \max_{\tau_{j}^{-} \in \tau^{-}}\left(\frac{1}{|\tau_j^-|}\sum_{t=0}^{|\tau_j^-|} \gamma^{t}r(s_{jt}^-, a_{jt}^-)\right)
\end{aligned}
\end{equation}
where $\tau_{i}^{+}=\{(s_{i0}^+, a_{i0}^+), (s_{i1}^+, a_{i1}^+), \dots, (s_{iT}^+, a_{iT}^+)\}$, $\tau_{j}^{-}=\{(s_{j0}^-, a_{j0}^-), \\ (s_{j1}^-, a_{j1}^-), \dots, (s_{jT}^-, a_{jT}^-)\}$.
\end{definition}

Using \TPE, the \critic can efficiently verify the effectiveness of the improved reward function. Specifically, if the designed reward function is order-preserving, trajectory preference sorting is considered to be reasonable and consistent with human expectations, making it more likely to improve the reward function. 
Since this process avoids redundant RL training and ground-truth rewards, preference feedback reduces both the cost of evaluating rewards and the dependence on prior knowledge. With the results of \TPE, the preference feedback includes the ratio of successful trajectory returns being higher than those of failed trajectories and provides details of two trajectories in the trajectory feedback format. Further details on preference feedback construction are presented in Appendix~\ref{app:introspection_prompt}.

\subsection{Reward Improvement}\label{sec:reward_improvement}

\designer of \ourmethod preserves a historical record of dialogues with the LLM, which includes system prompt, code generation prompt, and feedback prompts for each reward introspection. Through comprehensive context records, LLM gains an exhaustive understanding of the reward function enhancement process and is capable of immediately introspecting when negative optimization occurs.

\designer queries the LLM based on this feedback to improve the reward function. This process does not require direct LLM or human involvement, and only one or a minimal number of independent experiments is needed to achieve comparable or better results than earlier approaches. 

\subsection{Practical Implementation}\label{sec:practical_implementation}

Initially, \critic uses the code generated by \designer, which may not be optimal, to conduct RL training and collect trajectories. Process feedback and trajectory feedback are then constructed to allow \designer to refine the code. Each time \designer improves the reward function, \critic conducts \TPE on the updated code. If the reward function is order-preserving, RL training is performed, and process feedback and trajectory feedback are provided to \designer. \critic also records the generated trajectories for subsequent \TPE analysis. If the reward function is not order-preserving, it is considered unreasonable, and RL training is skipped to avoid unnecessary cost. Instead, preference feedback is provided, informing the LLM that the updated reward function may deviate from the optimal. The whole algorithm of \ourmethod with three types of feedback is shown in Appendix~\ref{app:full_algorithm}.

In implementation, we count the number of successful task trajectories whose average per-step return is greater than that of failed task trajectories to calculate the accuracy of the preference order. If the accuracy exceeds a threshold, the reward function is considered order-preserving. This approach provides flexibility in refining reward functions, allowing ample opportunities for modifications. During the first reward introspection, \critic conducts RL training directly and collects trajectories, as there are no initial trajectories for \TPE evaluation. Starting from the second reward introspection, the \TPE results determine whether RL training is necessary. \TPE significantly reduces unnecessary RL training and shortens the time spent on reward introspection. The results in Section~\ref{sec:results} show that alternating between process feedback, trajectory feedback, and preference feedback improves \designer's understanding of the environment, task goals, and reward functions, leading to the development of more effective reward functions.

\section{Experiments}\label{sec:experiments}

In this section, we conduct experiments to answer the following questions:
(\textbf{Q1}) How does \ourmethod perform on the evaluation tasks compared to the baselines?
(\textbf{Q2}) How does \ourmethod perform with respect to token efficiency?
(\textbf{Q3}) How sensitive is \ourmethod to its key parameters and components?
(\textbf{Q4}) How does \ourmethod perform in terms of execution error rate of the generated reward function code?
(\textbf{Q5}) How does \ourmethod improve the reward function with dynamic feedback?

\subsection{Experimental Setup}\label{sec:experimental_setup}

Following~\cite{Text2Reward}, we evaluate on $6$ robotic manipulation tasks of \MW~\cite{Meta-World} and $6$ robotic manipulation tasks of \MS~\cite{ManiSkill2}. Detailed descriptions of these tasks are provided in Appendix~\ref{app:task_details}.

\paragraph{Baselines.}
To evaluate the performance of \ourmethod, we compare it against the following baselines:
1) \textbf{Oracle}: The reward function is designed by human experts.
2) \textbf{\lr}~\cite{L2R}: \lr generates reward functions by combining human-defined sub-reward components.
3) \textbf{\tr}~\cite{Text2Reward}: \tr generates reward function code using zero-shot or few-shot methods by providing a textual description of the environment and task instructions.
4) \textbf{Eureka}~\cite{Eureka}: Eureka employs an evolutionary search strategy to improve the reward function by selecting the best-performing one from $16$ independent trials.

For fair comparison, all experiments are conducted in a zero-shot setting and \texttt{GPT-4-1106-preview}~\cite{GPT-4} API is used for \tr, \eureka, and \ourmethod. For \lr, we provide detailed task instructions for the Motion Descriptor and organize oracle reward components into separate reward APIs for the Reward Coder. This allows the LLM to perceive almost all relevant environment details, enabling \lr to perform similarly to Oracle. For other methods, we provide the same environment abstraction as \cite{Text2Reward}. Therefore, the range of variables that can be used when generating the reward function is the same, ensuring the fairness of the comparison. For \tr, we use both the human-improved code provided by~\cite{Text2Reward} and the generated code without human feedback (denoted as \textbf{Reproduce} in Section~\ref{sec:results}). We apply the same RL algorithm (PPO or SAC) and set the same hyperparameters as \tr~\cite{Text2Reward} across all methods. For \ourmethod, we set $\delta$ to $0.8$ and run $2$ iterations to refine the reward function code.
The experiments are conducted on a single NVIDIA RTX 3090 GPU with $8$ CPU cores.

\begin{figure*}[!t]
\centering
\includegraphics[width=0.8\textwidth]{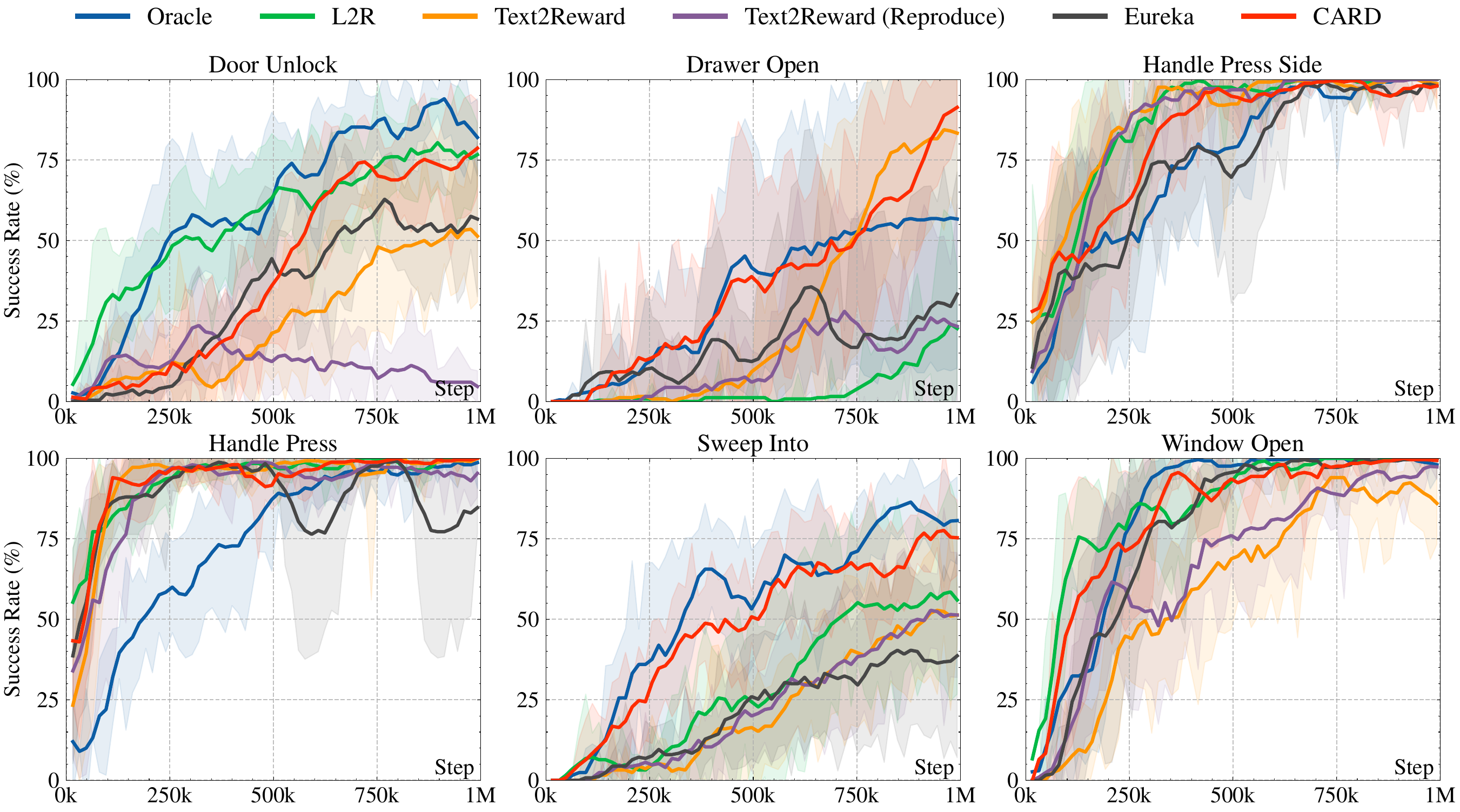}
\caption{Learning curves on six \textbf{\MW} tasks, measured by task success rate. The solid line represents the mean success rate, while the shaded regions correspond to the standard deviation, both calculated across five random seeds.}
\label{fig:metaworld_comparison}
\end{figure*}

\begin{figure*}[t]
\centering
\includegraphics[width=0.8\textwidth]{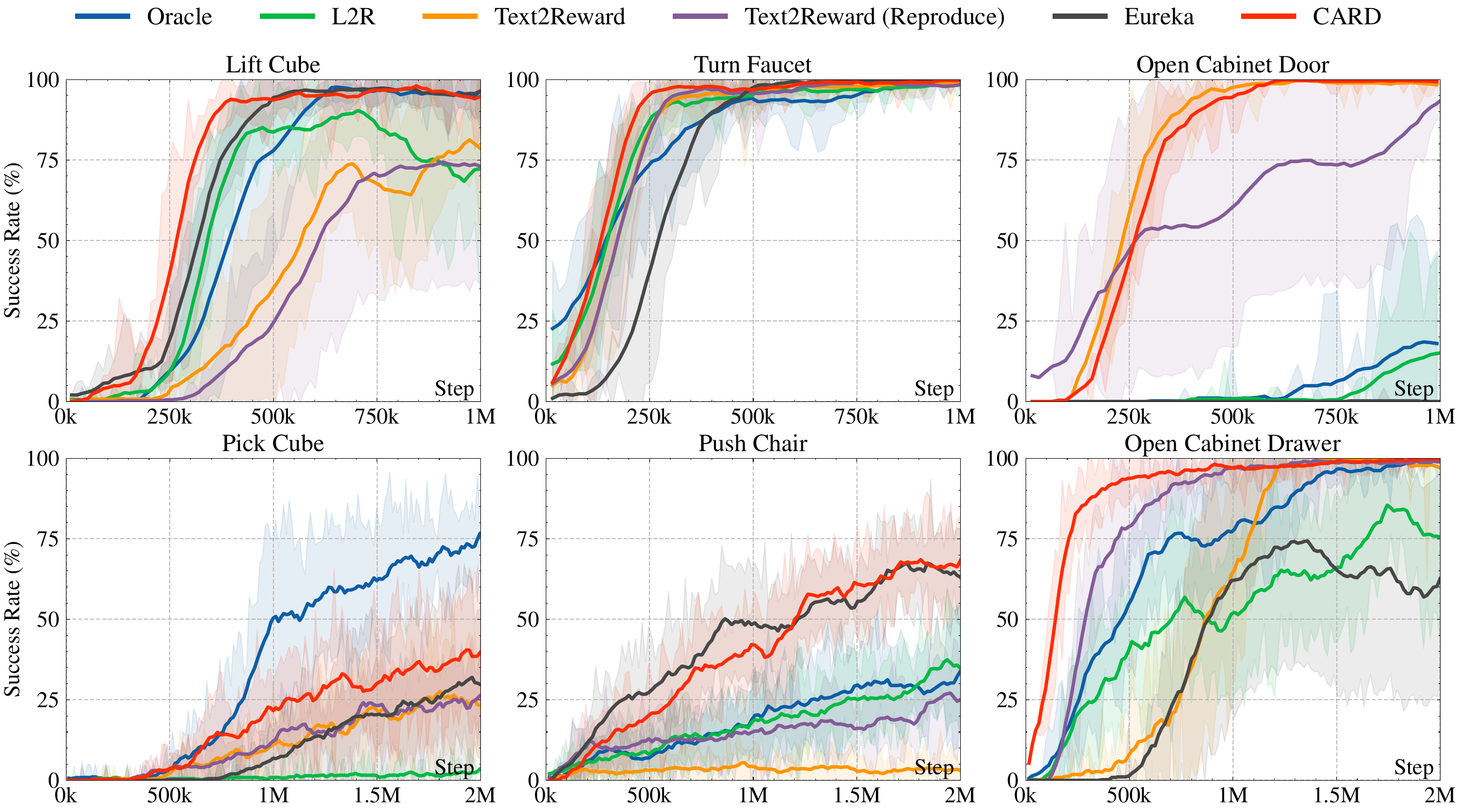}
\caption{Learning curves on six \textbf{\MS} tasks, measured by task success rate. The solid line represents the mean success rate, and the shaded areas denote the standard deviation, calculated across five random seeds.}
\label{fig:maniskill_comparison}
\end{figure*}

\subsection{Results (Q1)}\label{sec:results}

As shown in Figure~\ref{fig:metaworld_comparison} and Figure~\ref{fig:maniskill_comparison}, \ourmethod consistently achieves better or similar performance than other baselines on all tasks, while \ourmethod is also more cost-efficient. Notably, on the \textit{Sweep Into} task, \ourmethod demonstrates significant improvement compared to the baselines. On 3 out of 12 tasks, \ourmethod even surpasses oracle. For the remaining tasks, \ourmethod still performs comparably to oracle. We hypothesis that the reward function code generated by \ourmethod is highly effective given the available information, leaving minimal room for further improvement.

\subsection{Token Efficiency (Q2)}

To further compare the cost of LLM queries of different algorithms, we compute the total input and output tokens consumed. \lr queries the LLM once using the Motion Descriptor and once using the Reward Coder, so we sum the cost of both queries as the overall cost. \tr, without human feedback improvement, performs only one LLM query. \eureka generates initial reward code and refines it for $5$ iterations and each generation contains $16$ parallel sampling. Therefore, the token usage of \eureka is computed by the sum of $96$ queries. For \ourmethod, we measure the token consumption over initial generation and $2$ rounds of reward introspection, which requires $3$ LLM queries. In our experiments, token usage is calculated from 16 independent runs for both \lr and \tr, and we report the average values. Since both \eureka and \ourmethod involve iterations, we measure token consumption from one run due to cost limitations.

Table~\ref{tab:token_usage_simple} summarizes the average token consumption of different algorithms for LLM queries across all evaluation tasks. Although Text2Reward without iterative improvement consumes the least tokens, it still incurs additional token consumption and high labor costs when human feedback is introduced. \lr also consumes a small number of tokens because LLM only needs to select the reward API without requiring details of the environment settings. However, human efforts are needed to implement the reward API. \eureka consumes the most tokens, as it improves the reward function through parallel queries iteratively. \ourmethod strikes a balance between token consumption and performance, consuming fewer tokens than other iterative methods while performing better than non-iterative methods. Complete token consumption details for all algorithms are provided in Appendix~\ref{app:token_usage_details}.

\begin{table*}[!t]
\centering
\caption{Comparison of token usage among different methods averaged across all evaluation tasks.}
\resizebox{0.34\linewidth}{!}{
\begin{tabular}{lc}
\toprule
    \textbf{Method}
    & \textbf{Token Consumption}
    \\
\midrule
    \lr & $2310.7$ \\
    \tr & $1724.1$ \\
    \eureka & $662882.5$ \\
    \ourmethod & $14181.7$ \\
\bottomrule
\end{tabular}
}
\label{tab:token_usage_simple}
\end{table*}

\subsection{Ablation Study (Q3)}\label{sec:ablation}

\paragraph{Different number of iterations.}

\begin{figure*}[!ht]
\centering
\includegraphics[width=1\textwidth]{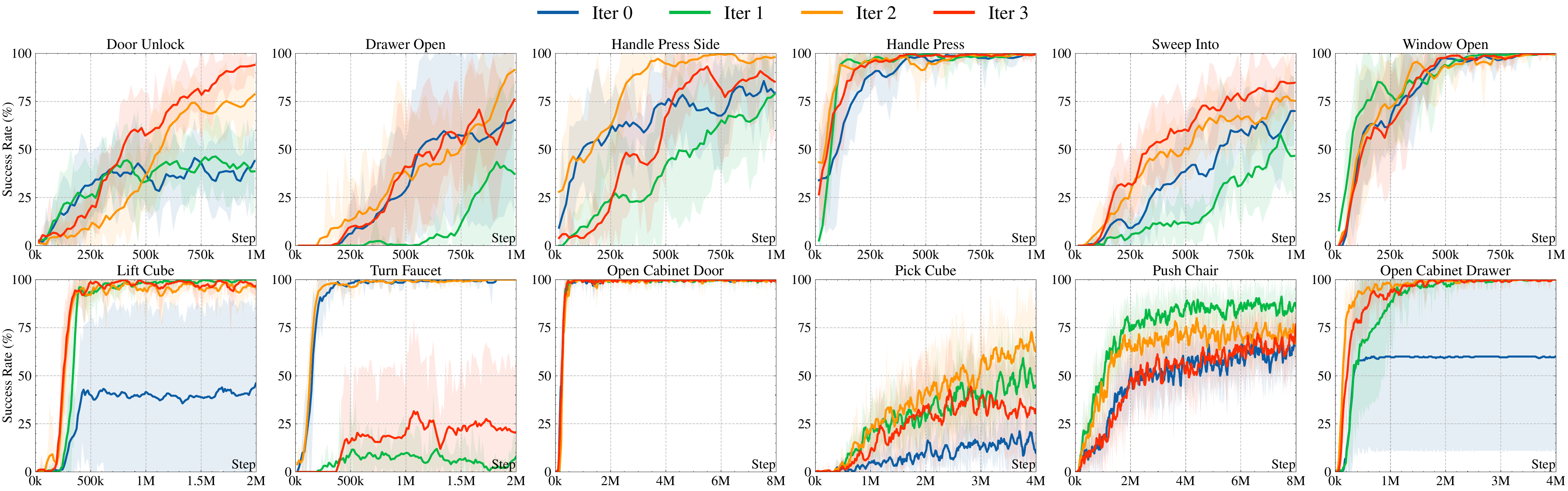}
\caption{Learning curves of \ourmethod on \textbf{\MW} and \textbf{\MS} tasks with different number of iterations. The solid line represents the mean success rate, and the shaded areas denote the standard deviation, calculated across five random seeds.}
\label{fig:ablation_iter}
\end{figure*}

To investigate how different number of iterations influences \ourmethod, we conduct experiments with iterations of $\{1,2,3\}$. The results in Figure~\ref{fig:ablation_iter} indicate that the quality of the reward function improves during the initial iterations and then gradually converges. In particular, on 7 out of 9 tasks, the results after the second iteration are either the best or comparable to those of the other iterations. Significant improvements are only observed in the third iteration for the more challenging tasks, such as \textit{Door Unlock} and \textit{Sweep Into}. However, on \textit{Handle Press Side} and \textit{Turn Faucet}, performance declines considerably after the third iteration. We hypothesis that after generating a well-shaped reward function, further refinement by the LLM may introduce unwanted changes, leading to a drop in performance. Therefore, for more difficult tasks, increasing the number of iterations is beneficial for fully refining the reward function, whereas for simpler tasks, limiting the number of refinement iterations may be more effective.

\paragraph{Different types of feedback.}

\begin{figure}[!ht]
\centering
\includegraphics[width=0.75\textwidth]{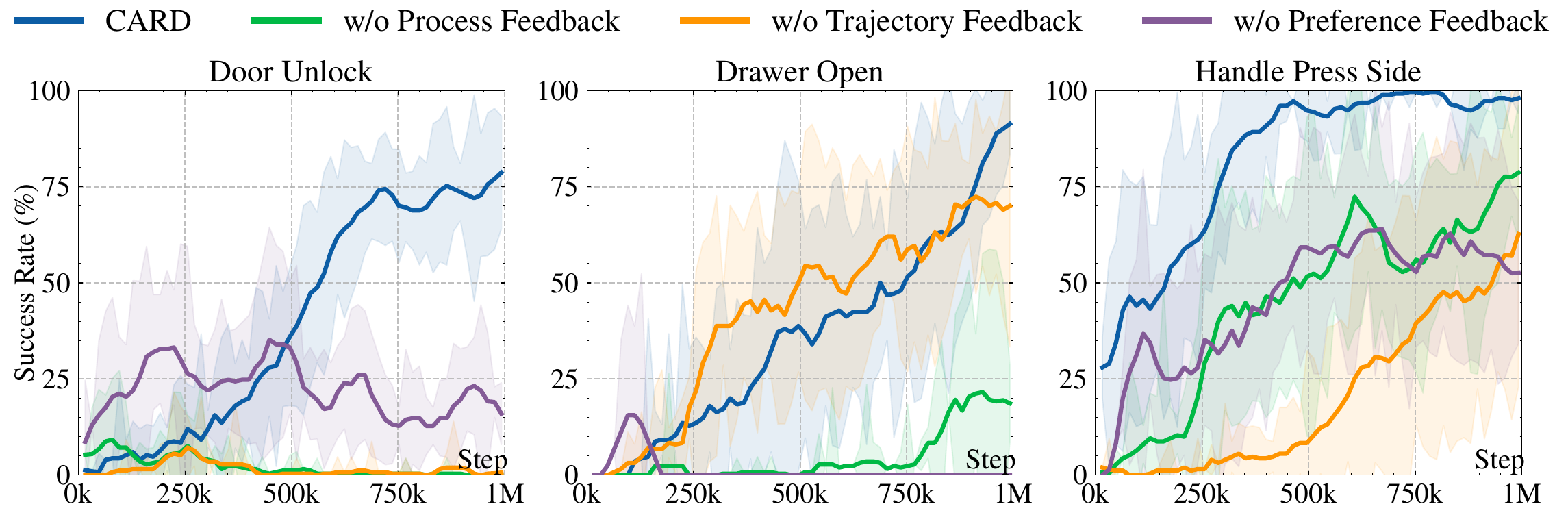}
\caption{Learning curves of \ourmethod on three \textbf{\MW} tasks with different types of feedback. The solid line represents the mean success rate, and the shaded areas denote the standard deviation, calculated across five random seeds.}
\label{fig:metaworld_feedback_ablation}
\end{figure}

To analyze the impact of different types of feedback on \ourmethod, we conduct experiments on three tasks (\textit{Door Unlock}, \textit{Drawer Open} and \textit{Handle Press Side}) of \MW by removing process feedback, trajectory feedback and preference feedback, respectively. As shown in Figure~\ref{fig:metaworld_feedback_ablation}, \ourmethod achieves the best performance with all types of feedback and removing either feedback can cause the performance to drop significantly.

\paragraph{Different LLM APIs.}

To examine the effect of different LLM APIs on \ourmethod, we conducted experiments using \texttt{GPT-3.5-turbo-1106} on three tasks: \textit{Door Unlock}, \textit{Drawer Open}, and \textit{Handle Press Side} from \MW. The results were then compared with those generated using \texttt{GPT-4-1106-preview}. The results show that \texttt{GPT-4} outperforms \texttt{GPT-3.5} on all tasks. In particular, for \textit{Door Unlock} and \textit{Drawer Open}, the improvement with \texttt{GPT-4} is substantial, highlighting its strong capabilities in generating reward function code.

\begin{figure}[!ht]
\centering
\includegraphics[width=0.75\textwidth]{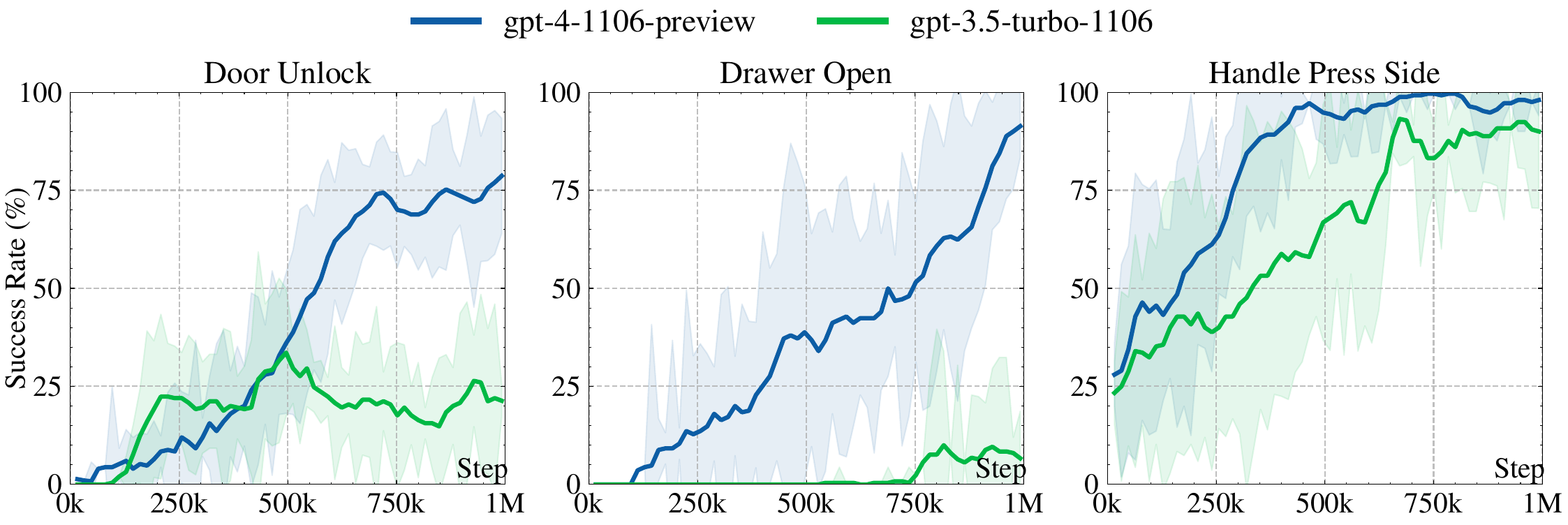}
\caption{Learning curves of \ourmethod on three \textbf{\MW} tasks using the reward functions generated by different LLM APIs. The solid line represents the mean success rate, and the shaded areas denote the standard deviation, calculated across five random seeds.}
\label{fig:metaworld_api_ablation}
\end{figure}

\subsection{Code Execution Error Rate (Q4)} \label{sec:execution_error_rate}

To compare the robustness of reward generation across different algorithms, we evaluate the execution error rate of the reward function code designed by LLM. For \lr and \tr, we independently run $16$ experiments and assess the execution error rate of the generated reward functions. For \eureka and \ourmethod, we calculate the results based on the first generated reward functions and the improved versions during the iteration process. We calculate the average over all tasks, as shown in Table~\ref{tab:execution_error_rate_simple}. The results show that \lr achieves the lowest code execution error rate. We attribute this to the fact that LLM only needs to determine the coefficients for different reward APIs, without having to implement detailed functions. On the other hand, \tr yields the highest error rate in code execution. By analyzing the erroneous code, we find that LLM often uses undefined variables or generates code that does not follow the correct template. The execution error rate of \ourmethod is lower than \eureka, demonstrating that our method is more token-efficient than \eureka without redundant parallel sampling. Detailed execution error rate results for LLM-designed reward functions across different algorithms are reported in Appendix~\ref{app:execution_error_rate_detailed}.

\begin{table}[!ht]
\centering
\caption{Comparison of execution error rate among different methods averaged across all evaluation tasks.}
\resizebox{0.45\linewidth}{!}{
\begin{tabular}{lc}
\toprule
    \textbf{Method} & \textbf{Code Execution Error Rate (\%)} \\
\midrule
    \lr & $6$ \\
    \tr & $48$ \\
    \eureka & $17$ \\
    \ourmethod & $12$ \\
\bottomrule
\end{tabular}
}
\label{tab:execution_error_rate_simple}
\end{table}

\subsection{Case Study (Q5)}\label{sec:case_study}

To understand how different types of feedback iteratively improve the reward function code through reward introspection, we present a case study on the \textit{Door Unlock} task.

\begin{figure*}[!ht]
\centering
\begin{tabular}{cc}
\subfloat[Comparison of reward functions before and after iteration $1$. \label{fig:abla_case_iter1}]{\includegraphics[width=0.48\linewidth]{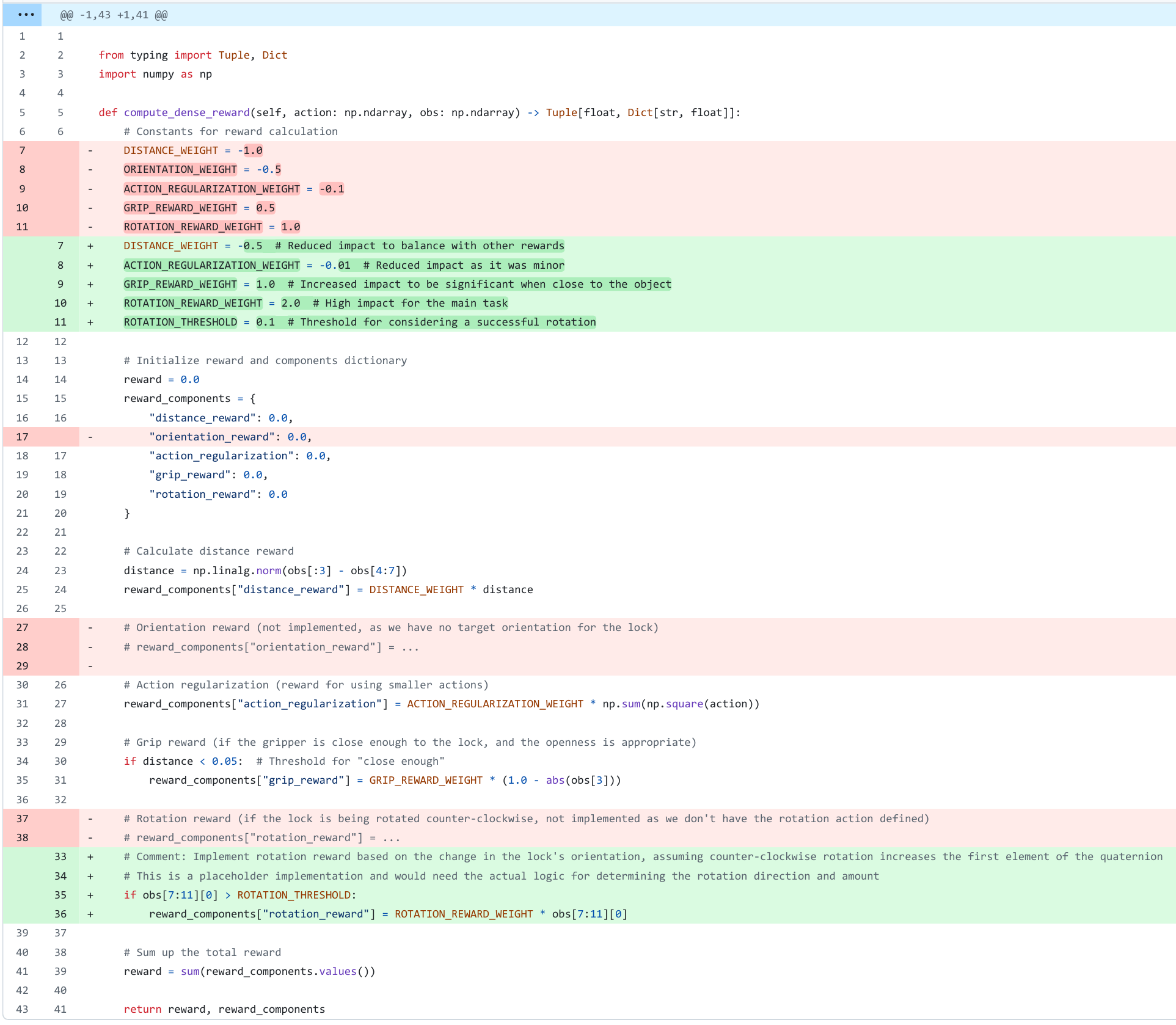}}
& \subfloat[Comparison of reward functions before and after iteration $2$. \label{fig:abla_case_iter2}]{\includegraphics[width=0.48\linewidth]{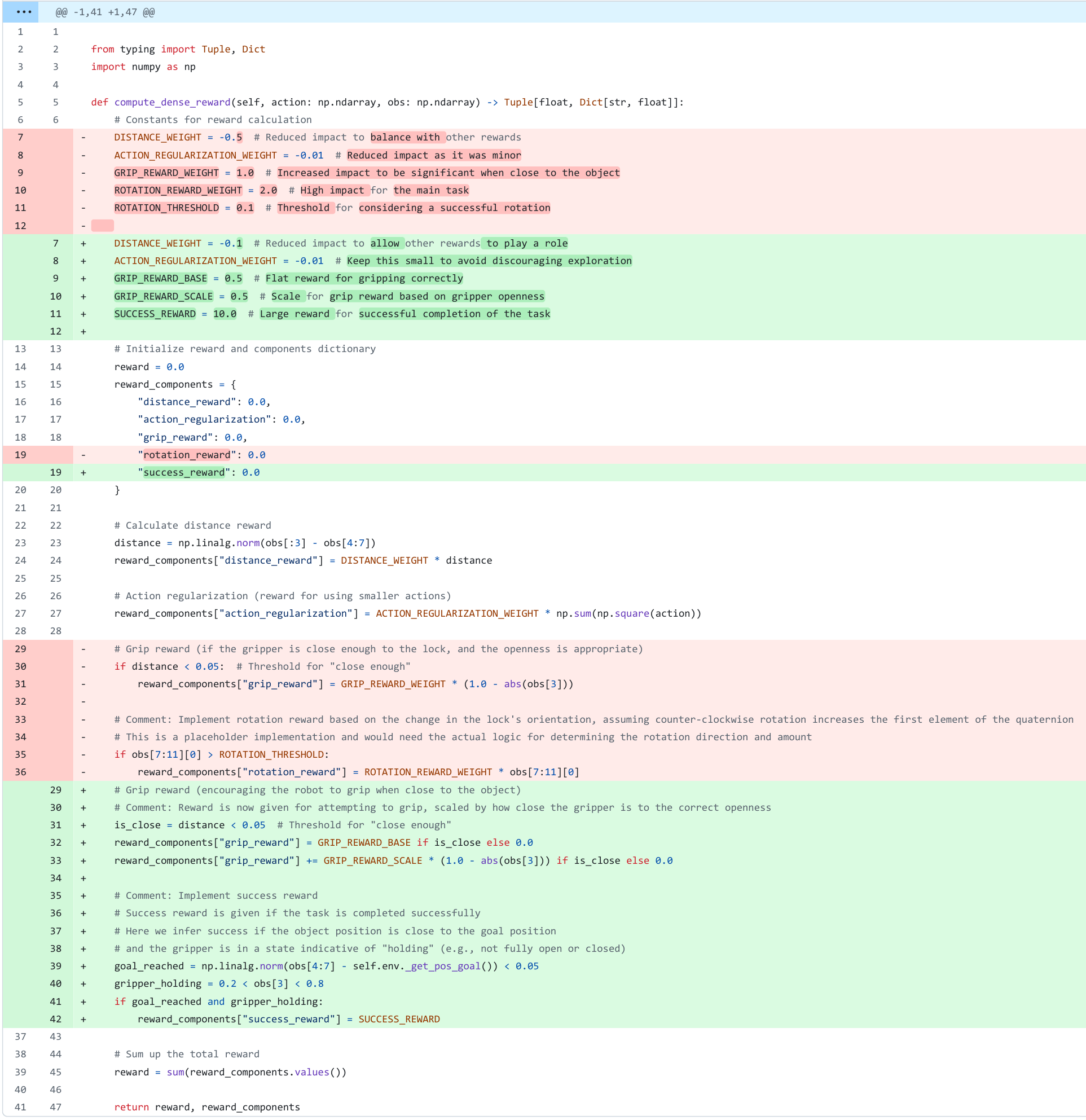}}
\end{tabular}
\caption{Changes in the reward function code when \ourmethod performs iteration $1$ (a) and iteration $2$ (b) reward introspection on \textit{Door Unlock}.}
\label{fig:abla_case}
\end{figure*}

\paragraph{Iteration 1.}

Figure~\ref{fig:abla_case_iter1} illustrates how \ourmethod improves the reward code during the first iteration of reward introspection. During this iteration, the \designer receives process feedback and trajectory feedback from RL training results. It can be seen that the LLM implements \verb|rotation_reward| based on the feedback and adjusts the weights for each reward component. Additionally, the \designer removes \verb|orientation_reward|, which is not implemented. From the results in Figure~\ref{fig:ablation_iter}, it can be observed that this refinement has no effect.

\paragraph{Iteration 2.}

In the next iteration, the LLM incorporates preference feedback, recognizing that successful trajectories should yield a higher return than failed ones. Figure~\ref{fig:abla_case_iter2} shows that the \designer introduces a \verb|success_reward|, providing a large reward for task completion. The \verb|grip_reward| is refined to include a base reward and a bonus to encourage the gripper to close. The \verb|rotation_reward| is removed, as it did not improve the results. After this iteration, the policy's success rate increases significantly, indicating that the reward function is now better aligned with trajectory preferences and task goals.

\section{Conclusion}

In this paper, we propose \ourmethod, a LLM-driven reward design framework that iteratively generates and refines reward function code without relying on redundant LLM queries or human feedback. The framework includes \designer, which is responsible for generating reward functions, and \critic, which evaluates these functions using dynamic feedback. In \ourmethod, we introduce Trajectory Preference Evaluation, which enables \critic to assess the reward functions without requiring RL training at every iteration, thus improving efficiency. The \critic provides multiple types of feedback, guiding the \designer in refining the reward code, creating a self-improving and automated process. Empirical results demonstrate that \ourmethod outperforms or matches the baselines on most tasks, and even surpasses the oracle on some tasks. Future work will explore extending \ourmethod to more complex environments and tasks.

\bibliographystyle{plain}
\bibliography{ref.bib}

\clearpage

\appendix

{\centering \LARGE \textbf{Appendix}\par}

\section{Algorithm}\label{app:full_algorithm}

The full algorithm of \ourmethod is presented in Algorithm~\ref{alg:algorithm}.

\begin{algorithm}[!htbp]
\caption{\ourmethod}
\label{alg:algorithm}
\begin{algorithmic}[1] 
\Require Environment description $M$, task instruction $l$, chat model in LLM $Coder$, offline text formatter $Evaluator$, system prompt $P_{\text{s}}$, instruction prompt $P_{\text{i}}$, feedback prompt $P_{\text{f}}$, preference evaluation function $F_p$, reinforcement learning training function $F_t$, Chat history $\mathcal{L}$, trajectories $\mathcal{T}$, iteration number $N$
\State $\mathcal{L}=\{P_{\text{s}}(M), P_{\text{i}}(l)\}$
\State // Reward Design
\State $R=\text{\designer}(\mathcal{L})$
\State $\text{result}=F_t(R)$
\State $\mathcal{T} \leftarrow \mathcal{T} \cup \{\text{result}\}$
\For{$N$ iteration}
\State // Reward Introspection
\State $\text{preference}=F_p(R)$
\If {$\text{preference}$}
\State $\text{result}=F_t(R)$
\State $\mathcal{T} \leftarrow \mathcal{T} \cup \{\text{result}\}$
\State $\mathcal{L} \leftarrow \mathcal{L} \cup \{P_{\text{f}}(\text{result})\}$
\Else
\State $\mathcal{L} \leftarrow \mathcal{L} \cup \{P_{\text{f}}(\text{preference})\}$
\EndIf
\State // Reward Improvement
\State $R=\text{\designer}(\mathcal{L})$
\EndFor
\Ensure Reward function code $R$
\end{algorithmic}
\end{algorithm}

\section{Experimental Details}\label{app:experimental_details}

In this section, we provide hyperparameter details used for reward design, reward introspection, and reward improvement.

\subsection{Task Details}\label{app:task_details}

\paragraph{\MW.}
In experimental evaluation, we use $6$ robotic manipulation tasks from \MW benchmark~\cite{Meta-World}, including \textit{Door Unlock}, \textit{Drawer Open}, \textit{Handle Press Side}, \textit{Handle Press}, \textit{Sweep Into} and \textit{Window Open}. We provide descriptions of each task as follows:
\begin{itemize}[leftmargin=20pt]
    \item \textit{Door Unlock}: The objective is to manipulate a robotic arm to unlock the door. The initial position of the door is random.
    \item \textit{Drawer Open}: The objective is to manipulate a robotic arm to open the drawer. The initial position of the drawer is random.
    \item \textit{Handle Press Side}: The objective is to manipulate a robotic arm to press the handle down sideways. The initial position of the handle is random.
    \item \textit{Handle Press}: The objective is to manipulate a robotic arm to press the handle. The initial position of the handle is random.
    \item \textit{Sweep Into}: The objective is to manipulate a robotic arm to sweep the puck into the hole. The initial position of the puck is random.
    \item \textit{Window Open}: The objective is to manipulate a robotic arm to open the window. The initial position of the window is random.
\end{itemize}
The instructions for generating reward code are listed in Table~\ref{tab:metaworld_tasks}.

\begin{table*}[!ht]
\centering
\caption{Instructions for generating reward code of each task in \MW.}
\begin{tabular}{ll}
\toprule
    \textbf{Task} & \textbf{Instruction} \\
\midrule
    Door Unlock & Unlock the door by rotating the lock counter-clockwise. \\
    Drawer Open & Open a drawer by its handle. \\
    Handle Press Side & Press a handle down sideways. \\
    Handle Press & Press a handle down. \\
    Sweep Into & Sweep a puck from the initial position into a hole. \\
    Window Open & Push and open a sliding window by its handle. \\
\bottomrule
\end{tabular}
\label{tab:metaworld_tasks}
\end{table*}

\paragraph{\MS.}
We utilize $6$ manipulation tasks from \MS~\cite{ManiSkill2}, including \textit{LiftCube}, \textit{PickCube}, \textit{TurnFaucet}, \textit{OpenCabinetDrawer}, \textit{OpenCabinetDoor} and \textit{PushChair}. The descriptions of tasks from \MS are as follows:
\begin{itemize}[leftmargin=20pt]
    \item \textit{LiftCube}: The objective is to lift the cube to the target position.
    \item \textit{PickCube}: The objective is to pick up the cube to the target position.
    \item \textit{TurnFaucet}: The objective is to turn on the handle of the faucet.
    \item \textit{OpenCabinetDrawer}: The objective is to manipulate the single-arm robot to open the drawer on the cabinet.
    \item \textit{OpenCabinetDoor}: The objective is to manipulate the single-arm robot to open the door on the cabinet.
    \item \textit{PushChair}: The objective is to manipulate the dual-arm robot to push the chair to the target position.
\end{itemize}
The instructions for generating reward code are listed in Table~\ref{tab:maniskill_tasks}.

\begin{table*}[!ht]
\centering
\caption{Instructions for generating reward code of each task in \MS.}
\begin{tabular}{ll}
\toprule
    \textbf{Task} & \textbf{Instruction} \\
\midrule
    LiftCube & Pick up cube A and lift it up by 0.2 meter. \\
    PickCube & Pick up cube A and move it to the 3D goal position. \\
    \multirow{2}{*}{TurnFaucet} & Turn on a faucet by rotating its handle. The task is finished \\
    & when qpos of faucet handle is larger than target qpos. \\
    \multirow{2}{*}{OpenCabinetDoor} & A single-arm mobile robot needs to open a cabinet door. The task is \\
    & finished when qpos of cabinet door is larger than target qpos. \\
    \multirow{2}{*}{OpenCabinetDrawer} & A single-arm mobile robot needs to open a cabinet drawer. The task is \\
    & finished when qpos of cabinet drawer is larger than target qpos. \\
    \multirow{2}{*}{PushChair} & A dual-arm mobile robot needs to push a swivel chair to a target \\
    & location on the ground and prevent it from falling over. \\
\bottomrule
\end{tabular}
\label{tab:maniskill_tasks}
\end{table*}

\subsection{Implementation Details}\label{app:implementation_details}

Since the \texttt{GPT-4-0314}~\cite{GPT-4} API used by~\cite{Text2Reward} has been deprecated~\footnote{\url{https://platform.openai.com/docs/models}}, we use \texttt{GPT-4-1106-preview}~\cite{GPT-4} API in all methods. And we use \texttt{GPT-3.5-turbo-1106} in Section~\ref{sec:ablation} to investigate the impact of different LLM APIs on \ourmethod. The temperature of sampling is set to $0.7$ in all tasks. For each generated reward function, we use dynamic execution to detect runtime errors. In the event of an error in the reward function code, we initiate a re-query to the LLM. The hyperparameter \textit{max\_try\_num} serves to prevent excessive costs associated with querying the LLM by either generating an executable code or ceasing the process after \textit{max\_try\_num} attempts. Throughout our experiments, we set \textit{max\_try\_num} to $10$.

We use the Proximal Policy Optimization (PPO)~\cite{schulman2017proximal} or Soft Actor-Critic (SAC)~\cite{haarnoja2018soft} algorithm to evaluate the reward function. Specifically, We use the open-source implementations of the above algorithms~\footnote{\url{https://github.com/DLR-RM/stable-baselines3}} by~\cite{raffin2021stable} and set the exact same hyperparameters as ~\cite{Text2Reward} for all methods. The hyperparameters on different benchmarks are shown in Table~\ref{tab:ppo_params} and Table~\ref{tab:sac_params}. We also provide the hyperparameter settings of \ourmethod, as shown in Table~\ref{tab:introspection_params}.

\begin{table*}[!ht]
\centering
\caption{Hyperparameters of PPO algorithm used in \textit{\LiftCube} and \textit{\PickCube} tasks of \MS.}
\begin{tabular}{ll}
\toprule
    \textbf{Hyperparameters} & \textbf{Value} \\
\midrule
    Learning rate & $3e^{-4}$ \\
    \# of layers & $2$ \\
    Hidden units per layer & $256$ \\
    \# of steps per update & $3200$ \\
    \# of epochs per update & $15$ \\
    Batch size & $400$ \\
    Discount factor $\gamma$ & $0.85$ \\
    Target KL divergence & $0.05$ \\
    \# of environments & $8$ \\
    GAE $\lambda$ & $0.95$ \\
    Clip range & $0.2$ \\
    Rollout steps per episode & $100$ \\
\bottomrule
\end{tabular}
\label{tab:ppo_params}
\end{table*}

\begin{table*}[!ht]
\centering
\caption{Hyperparameters of SAC algorithm used in \textit{\TurnFaucet}, \textit{\OpenCabinetDrawer}, \textit{\OpenCabinetDoor} and \textit{\PushChair} tasks of \MS and all \MW tasks.}
\begin{tabular}{ll}
\toprule
    \textbf{Hyperparameters} & \textbf{Value} \\
\midrule
    Learning rate & $3e^{-4}$ \\
    \# of layers & $3$ (\MW), $2$ (\MS) \\
    Hidden units per layer & $256$ \\
    Target update frequency & $2$ (\MW), $1$ (\MS) \\
    Train frequency & $1$ (\MW), $8$ (\MS) \\
    Soft update $\tau$ & $5e^{-3}$ \\
    Gradient steps & $1$ (\MW), $4$ (\MS) \\
    Learning starts & $4000$ \\
    Batch size & $512$ (\MW), $1024$ (\MS) \\
    Discount factor $\gamma$ & $0.99$ (\MW), $0.95$ (\MS) \\
    Initial temperature & $0.1$ (\MW), $0.2$ (\MS) \\
    \# of environments & $8$ \\
    Rollout steps per episode & $500$ (\MW), $200$ (\MS) \\
\bottomrule
\end{tabular}
\label{tab:sac_params}
\end{table*}

Each task is run on a single NVIDIA RTX 3090 GPU with $8$ CPU cores. To accelerate RL training processes, we leverage the vectorized environment acceleration provided by~\cite{raffin2021stable}. Consistent with the setting of~\cite{Text2Reward}, we use 8 parallel environments for training and 5 parallel environments for inference for all methods on all tasks. We demonstrate in Table~\ref{tab:time_cost} the time cost of RL training on distinct tasks, depending on the specified training steps and the aforementioned hyperparameter configurations. In our methodology, the actual time cost of each task is additionally influenced by the quantity of seeds and the number of introspection iterations.

\begin{table*}[!ht]
\centering
\caption{Hyperparameters of \ourmethod.}
\begin{tabular}{lc}
\toprule
    \textbf{Hyperparameters} & \textbf{Value} \\
\midrule
    \# of trajectories collected per iteration & $100$ \\
    Sampling interval for constructing process feedback & $100$ (\MW), $25$ (\MS) \\
    \# of sampling points for constructing trajectory feedback & $10$ \\
    Reward preference threshold $\delta$ & $0.8$ \\
\bottomrule
\end{tabular}
\label{tab:introspection_params}
\end{table*}

\begin{table*}[!ht]
\centering
\caption{Training steps and average time of each task.}
\begin{tabular}{llcc}
\toprule
    \textbf{Benchmark} & \textbf{Task} & \textbf{Steps} & \textbf{Average Time} \\
\midrule
    \MW & All & $1$M & $0.5$h \\
    \multirow{6}{*}{\MS} & LiftCube & $1$M & $0.5$h \\
    & TurnFaucet & $1$M & $0.9$h \\
    & OpenCabinetDoor & $1$M & $1.3$h \\
    & PickCube & $2$M & $1.4$h \\
    & PushChair & $2$M & $4.1$h \\
    & OpenCabinetDrawer & $2$M & $2.1$h \\
\bottomrule
\end{tabular}
\label{tab:time_cost}
\end{table*}

\subsection{Token Usage Details}\label{app:token_usage_details}
We report the detailed token consumption for all algorithms as shown in Table~\ref{tab:token_usage_detailed}.

\begin{table*}[!ht]
\centering
\caption{Detailed average token usage comparison among different algorithms.}
\resizebox{1\linewidth}{!}{
\begin{tabular}{lccccccc}
\toprule
    \textbf{Method}
    & \begin{tabular}[c]{c}\textbf{Door}\\ \textbf{Unlock}\end{tabular}
    & \begin{tabular}[c]{c}\textbf{Drawer}\\ \textbf{Open}\end{tabular}
    & \begin{tabular}[c]{c}\textbf{Handle}\\ \textbf{Press Side}\end{tabular}
    & \begin{tabular}[c]{c}\textbf{Handle}\\ \textbf{Press}\end{tabular}
    & \begin{tabular}[c]{c}\textbf{Sweep}\\ \textbf{Into}\end{tabular}
    & \begin{tabular}[c]{c}\textbf{Window}\\ \textbf{Open}\end{tabular}
    & \textbf{Average}
    \\
\midrule
    \lr & $2393$ & $2442$ & $2432$ & $2409$ & $2410$ & $2378$ & $2410.7$ \\
    \tr & $1153$ & $1153$ & $1158$ & $1142$ & $1151$ & $1105$ & $1143.7$ \\
    \eureka & $505648$ & $553311$ & $521594$ & $565891$ & $556693$ & $536400$ & $539922.8$ \\
    \ourmethod & $15023$ & $14586$ & $14536$ & $11568$ & $13069$ & $11324$ & $13351.0$ \\
\toprule
    \textbf{Method}
    & \begin{tabular}[c]{c}\textbf{Lift}\\ \textbf{Cube}\end{tabular}
    & \begin{tabular}[c]{c}\textbf{Open}\\ \textbf{CabinetDoor}\end{tabular}
    & \begin{tabular}[c]{c}\textbf{Open}\\ \textbf{CabinetDrawer}\end{tabular}
    & \begin{tabular}[c]{c}\textbf{Pick}\\ \textbf{Cube}\end{tabular}
    & \begin{tabular}[c]{c}\textbf{Push}\\ \textbf{Chair}\end{tabular}
    & \begin{tabular}[c]{c}\textbf{Turn}\\ \textbf{Faucet}\end{tabular}
    & \textbf{Average}
    \\
\midrule
    \lr & $2300$ & $2348$ & $2278$ & $2206$ & $2163$ & $1969$ & $2210.7$ \\
    \tr & $2321$ & $2344$ & $2328$ & $2263$ & $2277$ & $2294$ & $2304.5$ \\
    \eureka & $689889$ & $1013855$ & $568552$ & $766142$ & $755679$ & $920936$ & $785842.2$ \\
    \ourmethod & $14803$ & $15810$ & $16523$ & $13532$ & $14382$ & $15024$ & $15012.3$ \\
\bottomrule
\end{tabular}
}
\label{tab:token_usage_detailed}
\end{table*}

\subsection{Code Execution Error Rate}\label{app:execution_error_rate_detailed}

We report detailed execution error rate results for LLM-designed reward functions across different algorithms in Table~\ref{tab:execution_error_rate_detailed}.

\begin{table*}[!ht]
\centering
\caption{Detailed average execution error rate of generated code among different algorithms.}
\resizebox{1\linewidth}{!}{
\begin{tabular}{lccccccc}
\toprule
    \textbf{Method}
    & \begin{tabular}[c]{c}\textbf{Door}\\ \textbf{Unlock}\end{tabular}
    & \begin{tabular}[c]{c}\textbf{Drawer}\\ \textbf{Open}\end{tabular}
    & \begin{tabular}[c]{c}\textbf{Handle}\\ \textbf{Press Side}\end{tabular}
    & \begin{tabular}[c]{c}\textbf{Handle}\\ \textbf{Press}\end{tabular}
    & \begin{tabular}[c]{c}\textbf{Sweep}\\ \textbf{Into}\end{tabular}
    & \begin{tabular}[c]{c}\textbf{Window}\\ \textbf{Open}\end{tabular}
    & \textbf{Average}
    \\
\midrule
    \lr & $0.06\pm0.24$ & $0.06\pm0.24$ & $0.00\pm0.00$ & $0.00\pm0.00$ & $0.00\pm0.00$ & $0.00\pm0.00$ & $0.02$ \\
    \tr & $0.88\pm0.33$ & $0.31\pm0.46$ & $0.44\pm0.50$ & $0.63\pm0.48$ & $0.31\pm0.46$ & $0.31\pm0.46$ & $0.48$ \\
    \eureka & $0.32\pm0.23$ & $0.18\pm0.29$ & $0.23\pm0.25$ & $0.30\pm0.30$ & $0.11\pm0.12$ & $0.29\pm0.27$ & $0.24$ \\
    \ourmethod & $0.40\pm0.49$ & $0.00\pm0.00$ & $0.00\pm0.00$ & $0.00\pm0.00$ & $0.00\pm0.00$ & $0.00\pm0.00$ & $0.07$ \\
\toprule
    \textbf{Method}
    & \begin{tabular}[c]{c}\textbf{Lift}\\ \textbf{Cube}\end{tabular}
    & \begin{tabular}[c]{c}\textbf{Open}\\ \textbf{CabinetDoor}\end{tabular}
    & \begin{tabular}[c]{c}\textbf{Open}\\ \textbf{CabinetDrawer}\end{tabular}
    & \begin{tabular}[c]{c}\textbf{Pick}\\ \textbf{Cube}\end{tabular}
    & \begin{tabular}[c]{c}\textbf{Push}\\ \textbf{Chair}\end{tabular}
    & \begin{tabular}[c]{c}\textbf{Turn}\\ \textbf{Faucet}\end{tabular}
    & \textbf{Average}
    \\
\midrule
    \lr & $0.00\pm0.00$ & $0.00\pm0.00$ & $0.31\pm0.46$ & $0.13\pm0.33$ & $0.13\pm0.33$ & $0.06\pm0.24$ & $0.11$ \\
    \tr & $0.44\pm0.50$ & $0.25\pm0.43$ & $0.50\pm0.50$ & $0.44\pm0.50$ & $0.63\pm0.48$ & $0.56\pm0.50$ & $0.47$ \\
    \eureka & $0.07\pm0.10$ & $0.21\pm0.28$ & $0.09\pm0.16$ & $0.08\pm0.07$ & $0.05\pm0.09$ & $0.11\pm0.15$ & $0.10$ \\
    \ourmethod & $0.25\pm0.43$ & $0.25\pm0.43$ & $0.57\pm0.49$ & $0.00\pm0.00$ & $0.00\pm0.00$ & $0.00\pm0.00$ & $0.18$ \\
\bottomrule
\end{tabular}
}
\label{tab:execution_error_rate_detailed}
\end{table*}

\section{Prompt Details}
In this section, we provide prompt of \ourmethod, taking \MW tasks as an example.

\paragraph{System Prompt and Code Generation Prompt.}\label{app:system_and_generation_prompt}
The system prompt contains basic instruction of generating a reward function, and Python code that represents the Python class of the environment, the robot, and the rigid object. The system prompt follows~\cite{Text2Reward} and is shown in Table~\ref{tab:metaworld_system_prompt}. The task instruction, as shown in Table~\ref{tab:metaworld_tasks} and Table~\ref{tab:maniskill_tasks} is specified by fulfilling \texttt{\{instruction\}}.

\begin{table*}[!ht]
\centering
\caption{System prompt and code generation prompt of \MW tasks.}
\begin{tcolorbox}
\begin{lstlisting}[breaklines]
You are an expert in robotics, reinforcement learning and code generation.
We are going to use a robot arm to complete given tasks. The action space of the robot is a normalized `Box(-1, 1, (4,), float32)`.

Now I want you to help me write a reward function of reinforcement learning.
Typically, the reward function of a manipulation task is consisted of these following parts (some part is optional, so only include it if really necessary):
1. the distance between robot's gripper and our target object
2. difference between current state of object and its goal state
3. regularization of the robot's action
...

class BaseEnv(gym.env):
    self.robot : Robot # the robot in the environment
    self.obj1 : RigidObject # the first object in the environment
    self.obj2 : RigidObject # the second object in the environment, **if any**
    self.goal_position : np.ndarray[(3,)] # indicate the 3D position of the goal

class Robot:
    self.ee_position : np.ndarray[(3,)] # indicate the 3D position of the end-effector
    self.gripper_openness : float # a normalized measurement of how open the gripper is, range in [-1, 1]

class RigidObject:
    self.position : np.ndarray[(3,)] # indicate the 3D position of the rigid object
    self.quaternion : np.ndarray[(4,)] # indicate the quaternion of the rigid object

You are allowed to use any existing python package if applicable. But only use these packages when it's really necessary.

I want it to fulfill the following task: {instruction}
1. Please think step by step and tell me what does this task mean;
2. Then write a function that strictly follows the following format, and the return value of the function strictly follows the Type Annotation.
`
def compute_dense_reward(self, action: np.ndarray, obs: np.ndarray) -> Tuple[float, Dict[str, float]]:
    ...
    return reward, {...}
`
The output of the reward function should consist of two items:
    (1) the total reward,
    (2) a dictionary of each individual reward component.
The code output should be formatted as a python code string: "```python ... ```". Just the function body is fine.
3. Do not invent any variable or attribute that is not given.
4. When you writing code, you can also add some comments as your thought.
\end{lstlisting}
\end{tcolorbox}
\label{tab:metaworld_system_prompt}
\end{table*}

\paragraph{Introspection Prompt.}\label{app:introspection_prompt}
To prompt LLM to iteratively refine the reward code, we design introspection prompt, as shown in Table~\ref{tab:metaworld_introspection_prompt}. The introspection prompt contains three types of feedback: progress feedback, trajectory feedback and preference feedback, as described in Section~\ref{sec:reward_introspection}. 
Process feedback includes the average evaluation results of the agent, such as reward, episode length, success rate and each reward term. Trajectory feedback contains information of the evaluation results after RL training, including step-wise information of the trajectories with highest and lowest return.
Preference feedback provides the information of sorting the reward of successful trajectories and failed trajectories. If the generated reward code passes \TPE, the process feedback and trajectory feedback fulfills \texttt{\{inference\_results\}} in the introspection prompt, as shown in Table~\ref{tab:iterative_process_on_liftcube_1}.
If the generated code does not pass \TPE, the preference feedback will be provided, as shown in Table~\ref{tab:iterative_process_on_liftcube_2} and Table~\ref{tab:iterative_process_on_liftcube_3}.

\begin{table*}[!ht]
\centering
\caption{Introspection prompt of \MW tasks.}
\begin{tcolorbox}
\begin{lstlisting}[breaklines]
I use the reward function you wrote to train the RL agent. The feedback is as follows:
{inference_results}
1.Please carefully analyze the above feedback step by step, tell me your analysis results.
2.Consider how to enhance the reward function in order to increase the sample efficiency of RL and improve the task success rate. Provide a new, improved reward function that can better solve the task. Some helpful tips for analyzing the feedback:
    (1) If the success rates are always near zero, then you must rewrite the entire reward function
    (2) If the values for a certain reward component are near identical throughout, then this means RL is not able to optimize this component as it is written. You may consider
        (a) Changing its scale or the value of its temperature parameter
        (b) Re-writing the reward component 
        (c) Discarding the reward component
    (3) If some reward components' magnitude is significantly larger, then you must re-scale its value to a proper range
3. Give the content and reason of the modification in the form of comments before the modified position.
4. Do not invent any variable or attribute that is not given. Don't assume you can use other information.
\end{lstlisting}
\end{tcolorbox}
\label{tab:metaworld_introspection_prompt}
\end{table*}

\section{Examples of Reward Function and Reward Introspection}

\subsection{Reward Function Samples}
In this section, to help better understand the introspection capability of \ourmethod and the differences between our method and \tr, we select a challenging task from \MW and \MS respectively, and provide the reward functions of the two methods in the zero-shot setting.

Samples of the reward function designed by \ourmethod (after two rounds of iterations) and \tr on \MW \textit{Door Unlock} task are shown in Table~\ref{tab:ourmethod_code_doorunlock} and Table~\ref{tab:text2reward_code_doorunlock}, respectively. Samples of the reward function designed by \ourmethod (after two rounds of iterations) and \tr on \MS \textit{Lift Cube} task are shown in Table~\ref{tab:ourmethod_code_liftcube} and Table~\ref{tab:text2reward_code_liftcube}, respectively.

\begin{table*}[!ht]
\centering
\caption{Sample of the reward function designed by \ourmethod (after two rounds of iterations) on \MW \textit{Door Unlock} task.}
\begin{tcolorbox}
\begin{lstlisting}[language=python,breaklines]
from typing import Tuple, Dict
import numpy as np

def compute_dense_reward(self, action: np.ndarray, obs: np.ndarray) -> Tuple[float, Dict[str, float]]:
    # Comment: Reward constants should incentivize successful behavior more clearly
    DISTANCE_WEIGHT = -0.1  # Reduced impact to allow other rewards to play a role
    ACTION_REGULARIZATION_WEIGHT = -0.01  # Keep this small to avoid discouraging exploration
    GRIP_REWARD_BASE = 0.5  # Flat reward for gripping correctly
    GRIP_REWARD_SCALE = 0.5  # Scale for grip reward based on gripper openness
    SUCCESS_REWARD = 10.0  # Large reward for successful completion of the task

    # Initialize reward and components dictionary
    reward = 0.0
    reward_components = {
        "distance_reward": 0.0,
        "action_regularization": 0.0,
        "grip_reward": 0.0,
        "success_reward": 0.0
    }

    # Calculate distance reward
    distance = np.linalg.norm(obs[:3] - obs[4:7])
    reward_components["distance_reward"] = DISTANCE_WEIGHT * distance
    
    # Action regularization (reward for using smaller actions)
    reward_components["action_regularization"] = ACTION_REGULARIZATION_WEIGHT * np.sum(np.square(action))
    
    # Grip reward (encouraging the robot to grip when close to the object)
    # Comment: Reward is now given for attempting to grip, scaled by how close the gripper is to the correct openness
    is_close = distance < 0.05  # Threshold for "close enough"
    reward_components["grip_reward"] = GRIP_REWARD_BASE if is_close else 0.0
    reward_components["grip_reward"] += GRIP_REWARD_SCALE * (1.0 - abs(obs[3])) if is_close else 0.0

    # Comment: Implement success reward
    # Success reward is given if the task is completed successfully
    # Here we infer success if the object position is close to the goal position
    # and the gripper is in a state indicative of "holding" (e.g., not fully open or closed)
    goal_reached = np.linalg.norm(obs[4:7] - self.env._get_pos_goal()) < 0.05
    gripper_holding = 0.2 < obs[3] < 0.8
    if goal_reached and gripper_holding:
        reward_components["success_reward"] = SUCCESS_REWARD

    # Sum up the total reward
    reward = sum(reward_components.values())
    
    return reward, reward_components
\end{lstlisting}
\end{tcolorbox}
\label{tab:ourmethod_code_doorunlock}
\end{table*}

\begin{table*}[!ht]
\centering
\caption{Sample of the reward function designed by \tr on \MW \textit{Door Unlock} task.}
\begin{tcolorbox}
\begin{lstlisting}[language=python,breaklines]
import numpy as np
from scipy.spatial.distance import cdist

def compute_dense_reward(self, action, obs) -> float:
    # Define constants for reward tuning
    DISTANCE_WEIGHT = 1.0
    GOAL_REACHED_REWARD = 100.0
    ACTION_PENALTY = 0.1

    # Compute distance between robot's gripper and the lock
    distance = np.linalg.norm(obs[:3] - obs[4:7])

    # Compute difference between current state of object and its goal state
    goal_diff = np.linalg.norm(obs[4:7] - self.env._get_pos_goal())

    # Compute action regularization term
    action_penalty = ACTION_PENALTY * np.square(action).sum()

    # Check if the goal has been reached
    goal_reached = cdist(obs[4:7].reshape(1, -1), self.env._get_pos_goal().reshape(1, -1), 'cosine') < 0.01

    # Calculate reward
    reward = - DISTANCE_WEIGHT * distance - goal_diff - action_penalty
    if goal_reached:
        reward += GOAL_REACHED_REWARD

    return reward
\end{lstlisting}
\end{tcolorbox}
\label{tab:text2reward_code_doorunlock}
\end{table*}

\begin{table*}[!ht]
\centering
\caption{Sample of the reward function designed by \ourmethod (after two rounds of iterations) on \MS \textit{Lift Cube} task.}
\begin{tcolorbox}
\begin{lstlisting}[language=python,breaklines]
def compute_dense_reward(self, action: np.ndarray) -> Tuple[float, Dict[str, float]]:
    # Decrease the penalty for distance to encourage exploration without too much punishment.
    distance_weight = -1.0

    # Significantly increase the reward for lifting to make it much more significant.
    # This helps to clearly differentiate between successful and unsuccessful attempts.
    lifting_weight = 10.0

    # Slightly decrease the penalty for actions to prevent the agent from being too conservative.
    action_weight = -0.05

    # No change to the grasp reward as the agent seems to learn grasping well.
    grasp_weight = 0.5

    # Step 1: Calculate the distance between the robot's gripper and cube A
    ee_pose = self.tcp.pose
    cube_pose = self.obj.pose
    distance = np.linalg.norm(ee_pose.p - cube_pose.p)
    distance_reward = distance_weight * distance

    # Step 2: Check if the cube is lifted by 0.2 meters
    # The lifting reward is now proportional to the height but has a much larger weight.
    lifting_reward = lifting_weight * max(0.0, cube_pose.p[2] - 0.02)

    # Step 3: Apply regularization to the robot's action
    action_penalty = action_weight * np.linalg.norm(action)

    # Step 4: Check if the robot is grasping the cube
    grasp_reward = grasp_weight if self.agent.check_grasp(self.obj) else 0.0

    # Calculate total reward
    total_reward = distance_reward + lifting_reward + action_penalty + grasp_reward

    # Return the total reward and the dictionary of individual reward components
    return total_reward, {
        'distance_reward': distance_reward,
        'lifting_reward': lifting_reward,
        'action_penalty': action_penalty,
        'grasp_reward': grasp_reward,
    }
\end{lstlisting}
\end{tcolorbox}
\label{tab:ourmethod_code_liftcube}
\end{table*}

\begin{table*}[!ht]
\centering
\caption{Sample of the reward function designed by \tr on \MS \textit{Lift Cube} task.}
\begin{tcolorbox}
\begin{lstlisting}[language=python,breaklines]
import numpy as np
from scipy.spatial import distance

def compute_dense_reward(self, action) -> float:
    # Define reward weights
    weight_dist = 0.4
    weight_lift = 0.4
    weight_grasp = 0.2

    # Initialize reward
    reward = 0.0

    # Stage 1: Approach the cube
    ee_pos = self.tcp.pose.p
    cube_pos = self.obj.pose.p
    dist_to_cube = np.linalg.norm(ee_pos - cube_pos)
    reward_dist = -weight_dist * dist_to_cube
    
    # Stage 2: Grasp the cube
    grasp_success = self.agent.check_grasp(self.obj, max_angle=30)
    reward_grasp = weight_grasp * grasp_success

    # Stage 3: Lift the cube
    lift_amount = cube_pos[2] - self.goal_height
    reward_lift = -weight_lift * np.abs(lift_amount)

    # Total reward
    reward = reward_dist + reward_grasp + reward_lift

    # Stage 4: Maintain the cube at the goal height
    if self.obj.pose.p[2] >= self.goal_height:
        reward += 0.1 * (self.obj.pose.p[2] - self.goal_height)
        
    # Regularize the robot's action
    # We don't want robot to take very big action, so we add a negative reward here
    reward -= 0.01 * np.linalg.norm(action)

    return reward
\end{lstlisting}
\end{tcolorbox}
\label{tab:text2reward_code_liftcube}
\end{table*}

\subsection{Reward Introspection Results}
In order to clearly demonstrate the introspection process of \ourmethod, we select \textit{Lift Cube} task on \MS, and give the process of $3$ rounds of iterations on this tasks. The iterative process on \textit{Lift Cube} is shown in Table~\ref{tab:iterative_process_on_liftcube_0}, Table~\ref{tab:iterative_process_on_liftcube_1}, Table~\ref{tab:iterative_process_on_liftcube_2}, and Table~\ref{tab:iterative_process_on_liftcube_3}. The process demonstrates that \ourmethod iteratively refines the generated reward code based on multiple feedback information.

\begin{table*}[!ht]
\centering
\caption{The code first generated on \textit{Lift Cube}.}
\begin{tcolorbox}
\begin{lstlisting}[language=Python,breaklines]
def compute_dense_reward(self, action: np.ndarray) -> Tuple[float, Dict[str, float]]:
    # Initialize reward components
    distance_reward = 0.0
    lifting_reward = 0.0
    action_penalty = 0.0
    grasp_reward = 0.0
    total_reward = 0.0

    # Parameters for reward shaping
    distance_weight = -0.5  # Penalize the distance between gripper and cube A
    lifting_weight = 1.0    # Reward for lifting the cube to the desired height
    action_weight = -0.1    # Penalize large actions to encourage smooth movements
    grasp_weight = 0.5      # Encourage the robot to grasp the cube

    # Step 1: Calculate the distance between the robot's gripper and cube A
    ee_pose = self.tcp.pose
    cube_pose = self.obj.pose
    distance = np.linalg.norm(ee_pose.p - cube_pose.p)
    distance_reward = distance_weight * distance

    # Step 2: Check if the cube is lifted by 0.2 meters
    if cube_pose.p[2] >= self.goal_height:
        lifting_reward = lifting_weight

    # Step 3: Apply regularization to the robot's action
    action_penalty = action_weight * np.linalg.norm(action)

    # Step 4: Check if the robot is grasping the cube
    if self.agent.check_grasp(self.obj):
        grasp_reward = grasp_weight

    # Calculate total reward
    total_reward = distance_reward + lifting_reward + action_penalty + grasp_reward

    # Return the total reward and the dictionary of individual reward components
    return total_reward, {
        'distance_reward': distance_reward,
        'lifting_reward': lifting_reward,
        'action_penalty': action_penalty,
        'grasp_reward': grasp_reward,
    }
\end{lstlisting}
\end{tcolorbox}
\label{tab:iterative_process_on_liftcube_0}
\end{table*}

\begin{table*}[!ht]
\centering
\caption{The code generated after $1$ introspection iteration on \textit{Lift Cube}.}
\begin{tcolorbox}
\begin{lstlisting}[breaklines]
I use the reward function you wrote to train the RL agent. The feedback is as follows:
# Process Feedback:
1.Evaluation is performed after a certain number of training steps, and the average results of the episodes are as follows:
When step is 192000, reward is 48.52 (+-42.80), episode length is 200.00 (+-0.00), success rate is 0.00% (+-0.00%), distance_reward is -3.06 (+-1.43), lifting_reward is 0.00 (+-0.00), action_penalty is -9.85 (+-2.27), grasp_reward is 77.25 (+-38.68), 
......
# Trajectory Feedback
2.After training, we evaluated the model and the two trajectories with the highest and lowest return are as follows:
trajectories with the highest return 95.4080694151111 is:
When step is 1, reward is -0.22, distance_reward is -0.07, lifting_reward is 0.00, action_penalty is -0.16, grasp_reward is 0.00, 
......
trajectories with the lowest return 84.68022890798747 is:
When step is 1, reward is -0.18, distance_reward is -0.05, lifting_reward is 0.00, action_penalty is -0.13, grasp_reward is 0.00, 
......
# LLM Response
Analyzing the feedback:
1. The success rate is consistently 0%, indicating that the RL agent never successfully completes the task of lifting cube A by 0.2 meters within the given steps.
2. The grasp reward seems to be the only component contributing positively to the reward, and the agent appears to have learned to grasp the cube reliably.
3. The distance reward is negative in all cases and varies across evaluations. It seems that the agent does not consistently minimize the distance or that the distance calculation is not reflecting the desired behavior.
4. The action penalty seems to be functioning, penalizing larger actions, but it might be too small compared to the grasp reward, which could lead to the agent preferring to grasp without lifting.
......
def compute_dense_reward(self, action: np.ndarray) -> Tuple[float, Dict[str, float]]:
    # Parameters for reward shaping
    distance_weight = -2.0  # Increase the magnitude to encourage minimizing the distance more effectively
    lifting_weight = 2.0    # Increase the reward for lifting to make it more significant than the grasp reward
    action_weight = -0.1    # Keep the penalty for large actions
    grasp_weight = 0.5      # The agent seems to learn grasping well, no change needed
    # Step 1: Calculate the distance between the robot's gripper and cube A
    ee_pose = self.tcp.pose
    cube_pose = self.obj.pose
    distance = np.linalg.norm(ee_pose.p - cube_pose.p)
    distance_reward = distance_weight * distance
    # Step 2: Check if the cube is lifted by 0.2 meters
    # Modify the lifting reward to be proportional to the height achieved
    lifting_reward = lifting_weight * max(0.0, cube_pose.p[2] - 0.02)
    # Step 3: Apply regularization to the robot's action
    action_penalty = action_weight * np.linalg.norm(action)
    # Step 4: Check if the robot is grasping the cube
    grasp_reward = grasp_weight if self.agent.check_grasp(self.obj) else 0.0
    # Calculate total reward
    total_reward = distance_reward + lifting_reward + action_penalty + grasp_reward
    return total_reward, {
        'distance_reward': distance_reward,
        'lifting_reward': lifting_reward,
        'action_penalty': action_penalty,
        'grasp_reward': grasp_reward,
    }
\end{lstlisting}
\end{tcolorbox}
\label{tab:iterative_process_on_liftcube_1}
\end{table*}

\begin{table*}[!ht]
\centering
\caption{The code generated after $2$ introspection iteration on \textit{Lift Cube}.}
\begin{tcolorbox}
\begin{lstlisting}[breaklines]
......
# Preference Feedback:
We believe that the average reward for each step of most successful trajectories should be higher than that of failed trajectories......
For example, this is a trajectory where the agent successfully solved the task, with a return of -170.0785593882203, a length of 104, a average reward per step of -1.6353707633482721:
When step is 1, reward is -0.22, distance_reward is -0.07, lifting_reward is 0.00, action_penalty is -0.16, grasp_reward is 0.00......
However, the following is a trajectory that the agent failed to solve the task, with a return of -33.372598918676374, a length of 200, a average reward per step of -0.16686299459338186:
When step is 1, reward is -0.22, distance_reward is -0.07, lifting_reward is 0.00, action_penalty is -0.16, grasp_reward is 0.00......
# LLM Response
Analyzing the feedback:
1. The successful trajectory has an overall negative return due to the accumulation of negative rewards in the early steps, while the failed trajectory has a smaller negative return because it consistently receives a slightly positive reward throughout. This suggests that the negative components of the reward function (i.e., distance_reward and action_penalty) might be too punishing, or the positive rewards are not significant enough to outweigh the penalties.
2. In the successful trajectory, the lifting_reward is only received at the last steps, which is not enough to compensate for the accumulated penalties. This means that the reward for the main objective (lifting the cube) is not sufficiently high to differentiate successful from unsuccessful attempts.
3. The grasp_reward is consistently received in both successful and failed trajectories, indicating that the agent has learned to grasp the cube but not necessarily to lift it. This could be overshadowing the importance of the lifting reward.
Based on these observations, the reward function should be modified to:
- Increase the reward for the main task (lifting the cube).
- Decrease the penalties to a level that encourages exploration without too much punishment.
- Ensure that the reward function clearly distinguishes between successful and unsuccessful attempts.

def compute_dense_reward(self, action: np.ndarray) -> Tuple[float, Dict[str, float]]:
    # Decrease the penalty for distance to encourage exploration without too much punishment.
    distance_weight = -1.0
    # Significantly increase the reward for lifting to make it much more significant.
    # This helps to clearly differentiate between successful and unsuccessful attempts.
    lifting_weight = 10.0
    # Slightly decrease the penalty for actions to prevent the agent from being too conservative.
    action_weight = -0.05
    # No change to the grasp reward as the agent seems to learn grasping well.
    grasp_weight = 0.5
    # Step 1: Calculate the distance between the robot's gripper and cube A
    ee_pose = self.tcp.pose
    cube_pose = self.obj.pose
    distance = np.linalg.norm(ee_pose.p - cube_pose.p)
    distance_reward = distance_weight * distance
    # Step 2: Check if the cube is lifted by 0.2 meters
    # The lifting reward is now proportional to the height but has a much larger weight.
    lifting_reward = lifting_weight * max(0.0, cube_pose.p[2] - 0.02)
    # Step 3: Apply regularization to the robot's action
    action_penalty = action_weight * np.linalg.norm(action)
    # Step 4: Check if the robot is grasping the cube
    grasp_reward = grasp_weight if self.agent.check_grasp(self.obj) else 0.0
    # Calculate total reward
    total_reward = distance_reward + lifting_reward + action_penalty + grasp_reward
    return total_reward, {
        'distance_reward': distance_reward,
        'lifting_reward': lifting_reward,
        'action_penalty': action_penalty,
        'grasp_reward': grasp_reward,
    }
\end{lstlisting}
\end{tcolorbox}
\label{tab:iterative_process_on_liftcube_2}
\end{table*}

\begin{table*}[!ht]
\centering
\caption{The code generated after $3$ introspection iteration on \textit{Lift Cube}.}
\begin{tcolorbox}
\begin{lstlisting}[breaklines]
......
# Preference Feedback:
We believe that the average reward for each step of most successful trajectories should be higher than that of failed trajectories......
For example, this is a trajectory where the agent successfully solved the task, with a return of -85.03927969411015, a length of 104, a average reward per step of -0.8176853816741361: ......
However, the following is a trajectory that the agent failed to solve the task, with a return of -15.371678715571761, a length of 200, a average reward per step of -0.0768583935778588: ......
# LLM Response
Analyzing the feedback:
1. The successful trajectory has a negative average reward per step despite completing the task. This indicates that the penalties (distance_reward and action_penalty) are too severe or that the lifting_reward is not substantial enough to offset these penalties across all steps, especially early in the episode before the cube is lifted.
2. The failed trajectory has a smaller negative return because it consistently receives a positive grasp_reward that outweighs the penalties from distance_reward and action_penalty. This suggests that the grasp_reward may be too high relative to the penalties, allowing failed trajectories to accumulate higher rewards simply by grasping the cube without completing the task.
3. The agent receives a high lifting_reward only at the end of the successful trajectory. While it is important for the lifting_reward to be significant, it must also encourage incremental progress towards lifting the cube, or the agent may not be guided effectively through the intermediate steps of the task.

- The grasp_reward should be reduced to prevent it from dominating the reward signal.
- The lifting_reward should be fine-tuned to provide incremental rewards as the cube is lifted, rather than a large reward only at the goal height.
- The penalties should be adjusted to ensure that they encourage the correct behaviors without being too harsh.

def compute_dense_reward(self, action: np.ndarray) -> Tuple[float, Dict[str, float]]:
    # Reduce the penalty for distance to encourage exploration.
    distance_weight = -0.5  # Reduced severity as the agent was being penalized too much for distance.
    # Increase the reward for lifting. Provide incremental reward based on the height achieved,
    # not just on reaching the goal height. This will encourage progress towards lifting the cube.
    lifting_reward_multiplier = 5.0  # Increased to provide a stronger signal for lifting progress.
    # Reduce the penalty for actions to prevent the agent from being too conservative.
    action_weight = -0.02  # Reduced severity to encourage more exploratory actions.
    # Reduce the grasp reward as it was overshadowing other components.
    grasp_weight = 0.2  # Reduced to balance the overall reward structure.
    # Step 1: Calculate the distance between the robot's gripper and cube A
    ee_pose = self.tcp.pose
    cube_pose = self.obj.pose
    distance = np.linalg.norm(ee_pose.p - cube_pose.p)
    distance_reward = distance_weight * distance
    # Step 2: Check the height of the cube and provide incremental lifting reward
    lifting_progress = max(0.0, cube_pose.p[2] - 0.02)
    lifting_reward = lifting_reward_multiplier * lifting_progress
    # Step 3: Apply regularization to the robot's action
    action_penalty = action_weight * np.linalg.norm(action)
    # Step 4: Check if the robot is grasping the cube
    grasp_reward = grasp_weight if self.agent.check_grasp(self.obj) else 0.0
    # Calculate total reward
    total_reward = distance_reward + lifting_reward + action_penalty + grasp_reward
    # Return the total reward and the dictionary of individual reward components
    return total_reward, {
        'distance_reward': distance_reward,
        'lifting_reward': lifting_reward,
        'action_penalty': action_penalty,
        'grasp_reward': grasp_reward,
    }
\end{lstlisting}
\end{tcolorbox}
\label{tab:iterative_process_on_liftcube_3}
\end{table*}

\section{Limitations and Future Work}

Our research substantiates the efficacy of employing LLMs for the enhancement of dense functions in RL while eliminating human feedback. We design three distinct feedback categories, enabling the LLM to formulate reward functions composed of sub-rewards and to achieve introspection using the available information. Therefore, the reward improvement process exhibits high interpretability. Simultaneously, we remove invalid RL training and significantly improve the efficiency of introspection. 
However, as indicated in Table~\ref{tab:token_usage_detailed}, the token consumption of our method grows linearly with increasing iterations. Consequently, tasks necessitating extensive iterations to refine reward codes may incur substantial token usage. Moreover, our method presumes that the perception of environmental information is known a priori.
For future work, we consider utilizing Vision-Language Models (VLMs) to generate reward function with both vision and language information.

\end{document}